\documentclass{article}

\usepackage[accepted]{icml2021}

\usepackage[utf8]{inputenc} 
\usepackage[T1]{fontenc}    
\usepackage{hyperref}       
\usepackage{url}            
\usepackage{booktabs}       
\usepackage{amsfonts}       
\usepackage{nicefrac}       
\usepackage{microtype}      
\usepackage{graphicx}
\usepackage{subcaption}     
\usepackage{pgf}            
\usepackage{siunitx}        
\usepackage{enumitem}       
\usepackage{amsthm}         
\usepackage{wrapfig}        
\usepackage{arydshln}       
\usepackage{resizegather}
\usepackage{chngcntr}       
\usepackage{amssymb}        
\usepackage{placeins}       
\usepackage{etoolbox}       

\usepackage{amsmath,amsfonts,bm}
\usepackage{accents} 

\def\eqref#1{equation~\ref{#1}}

\def\1{\bm{1}}

\def\vone{{\bm{1}}}

\def\valpha{{\bm{\alpha}}}

\def\vb{{\bm{b}}}
\def\vc{{\bm{c}}}
\def\vd{{\bm{d}}}
\def\ve{{\bm{e}}}

\def\vh{{\bm{h}}}

\def\vk{{\bm{k}}}

\def\vp{{\bm{p}}}
\def\vq{{\bm{q}}}

\def\vs{{\bm{s}}}
\def\vt{{\bm{t}}}

\def\vx{{\bm{x}}}

\def\vz{{\bm{z}}}

\def\mA{{\bm{A}}}
\def\mB{{\bm{B}}}
\def\mC{{\bm{C}}}

\def\mI{{\bm{I}}}

\def\mK{{\bm{K}}}

\def\mM{{\bm{M}}}

\def\mP{{\bm{P}}}
\def\mQ{{\bm{Q}}}
\def\mR{{\bm{R}}}

\def\mU{{\bm{U}}}
\def\mV{{\bm{V}}}
\def\mW{{\bm{W}}}
\def\mX{{\bm{X}}}

\DeclareMathAlphabet{\mathsfit}{\encodingdefault}{\sfdefault}{m}{sl}
\SetMathAlphabet{\mathsfit}{bold}{\encodingdefault}{\sfdefault}{bx}{n}
\newcommand{\tens}[1]{\bm{\mathsfit{#1}}}

\def\tW{{\tens{W}}}

\def\gM{{\mathcal{M}}}

\def\gO{{\mathcal{O}}}

\def\sA{{\mathbb{A}}}

\def\sC{{\mathbb{C}}}

\newcommand{\E}{\mathbb{E}}

\newcommand{\R}{\mathbb{R}}

\newcommand{\KL}{D_{\mathrm{KL}}}

\DeclareMathOperator*{\argmax}{arg\,max}
\DeclareMathOperator*{\argmin}{arg\,min}

\newcommand{\unaryminus}{\scalebox{0.75}[1.0]{\( - \)}}

\newlength{\dtildeheight}

\newlength{\dcheckheight}

\usepackage[capitalize,nameinlink]{cleveref} 

\robustify\bfseries  
\newcommand\mcc[1]{\multicolumn{1}{c}{#1}}  
\crefname{section}{Sec.}{Sec.}
\crefname{appendix}{App.}{App.}
\crefname{definition}{Def.}{Defs.}

\newtheorem{definition}{Definition}
\newtheorem{theorem}{Theorem}
\newtheorem{proposition}{Proposition}

\newtheorem{lemma}{Lemma}

\pgfdeclarelayer{background}
\pgfdeclarelayer{foreground}
\pgfsetlayers{background,main,foreground}

\definecolor{uglyblue}{named}{blue}
\definecolor{uglygreen}{named}{green}
\definecolor{uglyred}{named}{red}
\definecolor{uglyyellow}{named}{yellow}

\definecolor{brewerpurple}{RGB}{117,112,179}
\definecolor{brewergreen}{RGB}{27,158,119}
\definecolor{brewerred}{RGB}{217,95,2}

\definecolor{blue}{RGB}{1, 115, 178}
\definecolor{orange}{RGB}{222, 143, 5}
\definecolor{green}{RGB}{2, 158, 115}
\definecolor{red}{RGB}{213, 94, 0}
\definecolor{pink}{RGB}{204, 120, 188}
\definecolor{yellow}{RGB}{236, 225, 51}

\makeatletter

\newcommand*{\@rowstyle}{}

\newcommand*{\rowstyle}[1]{
  \gdef\@rowstyle{#1}%
  \@rowstyle\ignorespaces%
}

\newcolumntype{=}{
  >{\gdef\@rowstyle{}}%
}

\newcolumntype{+}{
  >{\@rowstyle}%
}

\makeatother

\icmltitlerunning{Scalable Optimal Transport in High Dimensions for Graph Distances, Embedding Alignment, and More}

\begin{document}

\twocolumn[
\icmltitle{Scalable Optimal Transport in High Dimensions for Graph Distances, Embedding Alignment, and More}



\icmlsetsymbol{equal}{*}

\begin{icmlauthorlist}
\icmlauthor{Johannes Gasteiger}{tum}
\icmlauthor{Marten Lienen}{tum}
\icmlauthor{Stephan Günnemann}{tum}
\end{icmlauthorlist}

\icmlaffiliation{tum}{Technical University of Munich, Germany}

\icmlcorrespondingauthor{Johannes Gasteiger}{j.gasteiger@in.tum.de}

\icmlkeywords{Optimal Transport, Sinkhorn, Locality-Sensitive Hashing, Nyström, Graph Neural Networks, embedding alignment, graph distance}

\vskip 0.3in
]



\printAffiliationsAndNotice{}  

\begin{abstract}
The current best practice for computing optimal transport (OT) is via entropy regularization and Sinkhorn iterations. This algorithm runs in quadratic time as it requires the full pairwise cost matrix, which is prohibitively expensive for large sets of objects. In this work we propose two effective log-linear time approximations of the cost matrix: First, a sparse approximation based on locality-sensitive hashing (LSH) and, second, a Nyström approximation with LSH-based sparse corrections, which we call locally corrected Nyström (LCN). These approximations enable general log-linear time algorithms for entropy-regularized OT that perform well even for the complex, high-dimensional spaces common in deep learning. We analyse these approximations theoretically and evaluate them experimentally both directly and end-to-end as a component for real-world applications. Using our approximations for unsupervised word embedding alignment enables us to speed up a state-of-the-art method by a factor of 3 while also improving the accuracy by 3.1 percentage points without any additional model changes. For graph distance regression we propose the graph transport network (GTN), which combines graph neural networks (GNNs) with enhanced Sinkhorn. GTN outcompetes previous models by \SI{48}{\percent} and still scales log-linearly in the number of nodes.
\end{abstract}

\section{Introduction} \label{sec:intro}

Measuring the distance between two distributions or sets of objects is a central problem in machine learning. One common method of solving this is optimal transport (OT). OT is concerned with the problem of finding the transport plan for moving a source distribution (e.g.\ a pile of earth) to a sink distribution (e.g.\ a construction pit) with the cheapest cost w.r.t.\ some pointwise cost function (e.g.\ the Euclidean distance). The advantages of this method have been shown numerous times, e.g.\ in generative modelling \citep{arjovsky_wasserstein_2017,bousquet_optimal_2017,genevay_learning_2018}, loss functions \citep{frogner_learning_2015}, set matching \citep{wang_learning_2019}, or domain adaptation \citep{courty_joint_2017}. Motivated by this, many different methods for accelerating OT have been proposed in recent years \citep{indyk_fast_2003,papadakis_optimal_2014,backurs_scalable_2020}. However, most of these approaches are specialized methods that do not generalize to modern deep learning models, which rely on dynamically changing high-dimensional embeddings.

In this work we make OT computation for high-dimensional point sets more scalable by introducing two fast and accurate approximations of entropy-regularized optimal transport: Sparse Sinkhorn and LCN-Sinkhorn, the latter relying on our novel locally corrected Nyström (LCN) method. Sparse Sinkhorn uses a sparse cost matrix to leverage the fact that in entropy-regularized OT (also known as the Sinkhorn distance) \citep{cuturi_sinkhorn_2013} often only each point's nearest neighbors influence the result. LCN-Sinkhorn extends this approach by leveraging LCN, a general similarity matrix approximation that fuses local (sparse) and global (low-rank) approximations, allowing us to simultaneously capture interactions between both close and far points. LCN-Sinkhorn thus fuses sparse Sinkhorn and Nyström-Sinkhorn \citep{altschuler_massively_2019}. Both sparse Sinkhorn and LCN-Sinkhorn run in log-linear time.

\begin{figure}
    \centering
    \input{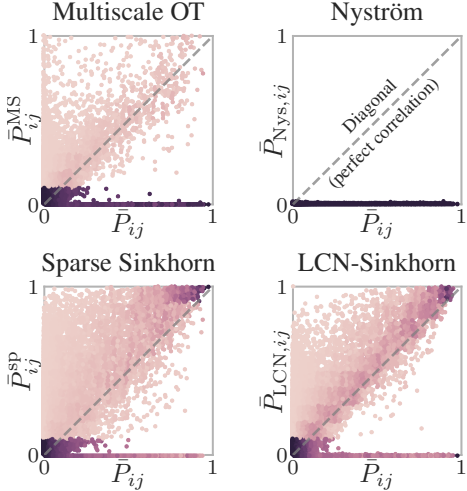}
    \caption{The proposed methods (sparse and LCN-Sinkhorn) show a clear correlation with the full Sinkhorn transport plan, as opposed to previous methods. Entries of approximations  (y-axis) and full Sinkhorn (x-axis) for pre-aligned word embeddings (EN-DE). Color denotes density.}
    \label{fig:ot-plan}
\end{figure}

We theoretically analyze these approximations and show that sparse corrections can lead to significant improvements over the Nyström approximation. We furthermore validate these approximations by showing that they are able to reproduce both the Sinkhorn distance and transport plan significantly better than previous methods across a wide range of regularization parameters and computational budgets (as e.g.\ demonstrated in \cref{fig:ot-plan}). We then show the impact of these improvements by employing Sinkhorn approximations end-to-end in two high-impact machine learning tasks. First, we incorporate them into Wasserstein Procrustes for word embedding alignment \citep{grave_unsupervised_2019}. Without any further model changes LCN-Sinkhorn improves upon the original method's accuracy by 3.1 percentage points using a third of the training time. Second, we develop the graph transport network (GTN), which combines graph neural networks (GNNs) with optimal transport for graph distance regression, and further improve it via learnable unbalanced OT and multi-head OT. GTN with LCN-Sinkhorn is the first model that both overcomes the bottleneck of using a single embedding per graph and scales log-linearly in the number of nodes. Our implementation is available online.\footnote{ \url{https://www.daml.in.tum.de/lcn}}
In summary, our paper's main contributions are:
\setlist{nolistsep}
\begin{itemize}[leftmargin=*,itemsep=2pt]
    \item \textbf{Locally Corrected Nyström (LCN)}, a flexible log-linear time approximation for similarity matrices, merging local (sparse) and global (low-rank) approximations.
    \item Entropy-regularized optimal transport (a.k.a.\ Sinkhorn distance) with log-linear runtime via \textbf{sparse Sinkhorn} and \textbf{LCN-Sinkhorn}. These are the first log-linear approximations that are stable enough to substitute full entropy-regularized OT in models using high-dimensional spaces.
    \item The \textbf{graph transport network (GTN)}, a siamese GNN using multi-head unbalanced LCN-Sinkhorn. GTN both sets the state of the art on graph distance regression and still scales log-linearly in the number of nodes.
\end{itemize}

\section{Entropy-regularized optimal transport}

This work focuses on optimal transport between two discrete sets of points. We use entropy regularization, which enables fast computation and often performs better than regular OT \citep{cuturi_sinkhorn_2013}. Formally, given two categorical distributions modelled via the vectors $\vp \in \R^{n}$ and $\vq \in \R^{m}$ supported on two sets of points $X_\text{p} = \{ \vx_{\text{p}1}, \dots, \vx_{\text{p}n} \}$ and $X_\text{q} = \{ \vx_{\text{q}1}, \dots, \vx_{\text{q}m} \}$ in $\R^d$ and the cost function $c: \R^d \times \R^d \rightarrow \R$ (e.g.\ the $L_2$ distance) giving rise to the cost matrix $\mC_{ij} = c(\vx_{\text{p}i}, \vx_{\text{q}i})$ we aim to find the Sinkhorn distance $d_c^\lambda$ and the associated optimal transport plan $\bar{\mP}$ \citep{cuturi_sinkhorn_2013}
\begin{equation}
\begin{gathered}
    \bar{\mP}=\argmin_{\mP} \langle \mP, \mC \rangle_\text{F} - \lambda H(\mP),\\
    d_c^\lambda = \langle \mP, \mC \rangle_\text{F} - \lambda H(\mP),\\
    \text{s.t.} \quad \mP \vone_m = \vp, \  \mP^T \vone_n = \vq,
\end{gathered}
\label{eq:entropy-ot}
\end{equation}
with the Frobenius inner product $\langle ., . \rangle_\text{F}$ and the entropy $H(\mP) = - \sum_{i=1}^n \sum_{j=1}^m \mP_{ij} \log \mP_{ij}$. Note that $d_c^\lambda$ includes the entropy and can thus be negative, while \citet{cuturi_sinkhorn_2013} originally used $d_{\text{Cuturi},c}^{1/\lambda} = \langle\bar{\mP}, \mC\rangle_\text{F}$. This optimization problem can be solved by finding the vectors $\bar{\vs}$ and $\bar{\vt}$ that normalize the columns and rows of the matrix $\bar{\mP} = \operatorname{diag}(\bar{\vs}) \mK \operatorname{diag}(\bar{\vt})$ with the similarity matrix $\mK_{ij} = e^{-\frac{\mC_{ij}}{\lambda}}$, so that $\bar{\mP} \vone_m = \vp$ and $\bar{\mP}^T \vone_n = \vq$. We can achieve this via the Sinkhorn algorithm, which initializes the normalization vectors as $\vs^{(1)} = \vone_n$ and $\vt^{(1)} = \vone_m$ and updates them alternatingly via \citep{sinkhorn_concerning_1967}
\begin{equation}
    \vs^{(i)} = \vp \oslash (\mK \vt^{(i-1)}),
    \quad
    \vt^{(i)} = \vq \oslash (\mK^T \vs^{(i)})
\label{eq:sinkhorn}
\end{equation}
until convergence, where  $\oslash$ denotes elementwise division.

\section{Sparse Sinkhorn} \label{sec:sparse-sinkhorn}

The Sinkhorn algorithm is faster than non-regularized EMD algorithms, which run in $\mathcal{O}(n^2m \log n \log(n \max(\mC)))$ \citep{tarjan_dynamic_1997}. However, its computational cost is still quadratic in time $\mathcal{O}(nm)$, which is prohibitively expensive for large $n$ and $m$. We overcome this by observing that the matrix $\mK$, and hence also $\bar{\mP}$, is negligibly small everywhere except at each point's closest neighbors because of the exponential used in $\mK$'s computation. We propose to leverage this by approximating $\mC$ via the sparse matrix $\mC^\text{sp}$, where
\begin{equation}
    \mC^\text{sp}_{ij} = \begin{cases}
    \mC_{ij} & \text{if $\vx_{\text{p}i}$ and $\vx_{\text{q}j}$ are ``near''},\\
    \infty & \text{otherwise}.
    \end{cases}
\end{equation}
$\mK^\text{sp}$ and $\bar{\mP}^\text{sp}$ follow from the definitions of $\mK$ and $\bar{\mP}$. Finding ``near'' neighbors can be approximately solved via locality-sensitive hashing (LSH) on $X_\text{p} \cup X_\text{q}$.

\textbf{Locality-sensitive hashing.} LSH tries to filter ``near'' from ``far'' data points by putting them into different hash buckets. Points closer than a certain distance $r_1$ are put into the same bucket with probability at least $p_1$, while those beyond some distance $r_2 = c \cdot r_1$ with $c > 1$ are put into the same bucket with probability at most $p_2 \ll p_1$. There is a plethora of LSH methods for different metric spaces and their associated cost (similarity/distance) functions \citep{wang_hashing_2014,shrivastava_asymmetric_2014}, and we can use any of them. In this work we focus on cross-polytope LSH \citep{andoni_practical_2015} and $k$-means LSH \citep{pauleve_locality_2010} (see \cref{app:landmarks}). We can control the (average) number of neighbors via the number of hash buckets. This allows sparse Sinkhorn with LSH to scale log-linearly with the number of points, i.e.\ $\mathcal{O}(n \log n)$ for $n \approx m$ (see \cref{app:complexity} and \cref{app:runtimes}). Unfortunately, Sinkhorn with LSH can fail when e.g.\ the cost is evenly distributed or the matrix $\mK^\text{sp}$ does not have support (see \cref{app:limitations}). However, we can alleviate these limitations by fusing $\mK^\text{sp}$ with the Nyström method.

\section{Locally corrected Nyström and LCN-Sinkhorn} \label{sec:lcn-sinkhorn}

\textbf{Nyström method.} The Nyström method is a popular way of approximating similarity matrices that provides performance guarantees for many important tasks \citep{williams_using_2001,musco_recursive_2017}. It approximates a positive semi-definite (PSD) similarity matrix $\mK$ via its low-rank decomposition $\mK_\text{Nys} = \mU \mA^{-1} \mV$. Since the optimal decomposition via SVD is too expensive to compute, Nyström instead chooses a set of $l$ landmarks $L = \{\vx_{\text{l}1}, \dots, \vx_{\text{l}l} \}$ and obtains the matrices via $\mU_{ij} = k(\vx_{\text{p}i}, \vx_{\text{l}j})$, $\mA_{ij} = k(\vx_{\text{l}i}, \vx_{\text{l}j})$, and $\mV_{ij} = k(\vx_{\text{l}i}, \vx_{\text{q}j})$, where $k(\vx_1, \vx_2)$ is an arbitrary PSD kernel, e.g.\ $k(\vx_1, \vx_2) = e^{-\frac{c(\vx_1, \vx_2)}{\lambda}}$ for Sinkhorn. Common methods of choosing landmarks from $X_\text{p} \cup X_\text{q}$ are uniform and ridge leverage score (RLS) sampling. We instead focus on $k$-means Nyström and sampling via $k$-means++, which we found to be significantly faster than recursive RLS sampling \citep{zhang_improved_2008} and perform better than both uniform and RLS sampling (see \cref{app:landmarks}).

\textbf{Sparse vs. Nyström.} Exponential kernels like the one used for $\mK$ (e.g. the Gaussian kernel) typically correspond to a reproducing kernel Hilbert space that is infinitely dimensional. The resulting Gram matrix $\mK$ thus usually has full rank. A low-rank approximation like the Nyström method can therefore only account for its global structure and not the local structure around each point $\vx$. As such, it is ill-suited for any moderately low entropy regularization parameter, where the transport matrix $\bar{\mP}$ resembles a permutation matrix. Sparse Sinkhorn, on the other hand, cannot account for global structure and instead approximates all non-selected distances as infinity. It will hence fail if more than a handful of neighbors are required per point. These approximations are thus opposites of each other, and as such not competing but rather \emph{complementary} approaches.

\textbf{Locally corrected Nyström.} Since the entries in our sparse approximation are exact, we can directly fuse it with the Nyström approximation. For the indices of all non-zero values in the sparse approximation $\mK^\text{sp}$ we calculate the corresponding entries in the Nyström approximation, obtaining the sparse matrix $\mK^\text{sp}_\text{Nys}$. To obtain the locally corrected Nyström (LCN) approximation\footnote{LCN has an unrelated namesake on integrals, which uses high-order term to correct quadrature methods around singularities \citep{canino_numerical_1998}.} we subtract these entries from $\mK_\text{Nys}$ and replace them with their exact values, i.e.
\begin{equation}
    \mK_\text{LCN} = \mK_\text{Nys} - \mK^\text{sp}_\text{Nys} + \mK^\text{sp}
    = \mK_\text{Nys} + \mK^\text{sp}_\Delta.
\end{equation}
\textbf{LCN-Sinkhorn.} To obtain the approximate transport plan $\bar{\mP}_\text{LCN}$ we run the Sinkhorn algorithm with $\mK_\text{LCN}$ instead of $\mK$. However, we never fully instantiate $\mK_\text{LCN}$. Instead, we directly use the decomposition and calculate the matrix-vector product in \cref{eq:sinkhorn} as $\mK_\text{LCN} \vt = \mU (\mA^{-1} \mV \vt) + \mK^\text{sp}_\Delta \vt$, similarly to \citet{altschuler_massively_2019}. As a result we obtain the decomposed approximate OT plan $\bar{\mP}_\text{LCN} = \bar{\mP}_\text{Nys} + \bar{\mP}^\text{sp}_\Delta = \bar{\mP}_{U} \bar{\mP}_{W} + \bar{\mP}^\text{sp} - \bar{\mP}^\text{sp}_\text{Nys}$ and the approximate distance (using Lemma A from \citet{altschuler_massively_2019})
\begin{equation}
\begin{split}
    d_{\text{LCN},c}^\lambda =
    \lambda &(
    \vs^T \bar{\mP}_{U} \bar{\mP}_{W} \vone_m
    + \vone_n^T \bar{\mP}_{U} \bar{\mP}_{W} \vt\\
    &+ \vs^T \bar{\mP}^\text{sp}_\Delta \vone_m
    + \vone_n^T \bar{\mP}^\text{sp}_\Delta \vt
    ).
\label{eq:lcn-sinkhorn}
\end{split}
\end{equation}
This approximation scales log-linearly with dataset size (see \cref{app:complexity} and \cref{app:runtimes} for details). It allows us to smoothly move from Nyström-Sinkhorn to sparse Sinkhorn by varying the number of landmarks and neighbors. We can thus freely choose the optimal ``operating point'' based on the underlying problem and regularization parameter. We discuss the limitations of LCN-Sinkhorn in \cref{app:limitations}.

\section{Theoretical analysis} \label{sec:theory}

\textbf{Approximation error.} The main question we aim to answer in our theoretical analysis is what improvements to expect from adding sparse corrections to Nyström Sinkhorn. To do so, we first analyse approximations of $\mK$ in a uniform and a clustered data model. In these we use Nyström and LSH schemes that largely resemble $k$-means, as used in most of our experiments. Relevant proofs and notes for this section can be found in \cref{app:uniform_error,app:cluster_error,app:sinkhorn_error,app:convergence,app:derivatives}.

\begin{theorem}
\label{th:uniform_error}
Let $X_\textnormal{p}$ and $X_\textnormal{q}$ have $n$ samples that are uniformly distributed on a $d$-dimensional closed, locally Euclidean manifold. Let $\mC_{ij} = \| \vx_{\textnormal{p}i} - \vx_{\textnormal{q}j} \|_2$ and $\mK_{ij} = e^{-\mC_{ij} / \lambda}$. Let the landmarks be arranged a priori, with a minimum distance $2R$ between each other. Then the expected error of the Nyström approximation $\mK_\textnormal{Nys}$ between a point $\vx_{\textnormal{p}i}$ and its $k$th-nearest neighbor $\vx_{\textnormal{q}i_k}$ is
\upshape
\begin{equation}
\label{eq:uniform_error}
    \E[ \mK_{i, i_k} - \mK_{\text{Nys}, i,i_k} ] = \E[e^{-\delta_k/\lambda}] - \E[\mK_{\text{Nys}, i,i_k}],
\end{equation}
\itshape
with $\delta_k$ denoting the $k$th-nearest neighbor distance. With $\Gamma(., .)$ denoting the upper incomplete Gamma function the second term is bounded by
\upshape
\begin{equation}
    \E[\mK_{\text{Nys}, i,i_k}] \le \frac{d (\Gamma(d) - \Gamma(d, 2R/\lambda))}{(2R/\lambda)^d} + \mathcal{O}(e^{-2R/\lambda}).
\end{equation}
\itshape
\end{theorem}

\cref{eq:uniform_error} is therefore dominated by $\E[e^{-\delta_k/\lambda}]$ if $\delta_k \ll R$, which is a reasonable assumption given that $R$ only decreases slowly with the number of landmarks $l$ since $R \ge (\frac{(d/2)!}{l})^{1/d}\frac{1}{2\sqrt{\pi}}$ \citep{cohn_conceptual_2017}. In this case the approximation's largest error is the one associated with the point's nearest neighbor. LCN uses the exact result for these nearest neighbors and therefore removes the largest errors, providing significant benefits even for uniform data. For example, just removing the first neighbor's error we obtain a \SI{68}{\percent} decrease in the dominant first term ($d\!=\!32$, $\lambda\!=\!0.05$, $n\!=\!1000$). This is even more pronounced in clustered data.

\begin{theorem}
\label{th:cluster_error}
Let $X_\textnormal{p}, X_\textnormal{q} \subseteq \R^d$ be inside $c$ (shared) clusters. Let $r$ be the maximum $L_2$ distance of a point to its cluster center and $D$ the minimum distance between different cluster centers, with $r \ll D$. Let $\mC_{ij} = \| \vx_{\textnormal{p}i} - \vx_{\textnormal{q}j} \|_2$ and $\mK_{ij} = e^{-\mC_{ij} / \lambda}$. Let each LSH bucket used for the sparse approximation $\mK^\textnormal{sp}$ cover at least one cluster. Let $\mK_\textnormal{Nys}$ and $\mK_\textnormal{LCN}$ both use one landmark at each cluster center. Then the maximum possible error is
\upshape
\begin{align}
\label{eq:cluster_error_nys}
    \begin{split}
    \max_{\vx_{\textnormal{p}i}, \vx_{\textnormal{q}j}} \mK_{ij} - \mK_{\textnormal{Nys},i,j} = {}& \\
    1 - {}& e^{-2r/\lambda} -  \mathcal{O}(e^{-2(D-r)/\lambda}),
    \end{split}\\
    \max_{\vx_{\textnormal{p}i}, \vx_{\textnormal{q}j}} \mK_{ij} - \mK^\textnormal{sp}_{ij} = {}& e^{-(D-2r)/\lambda},\\
    \begin{split}
    \max_{\vx_{\textnormal{p}i}, \vx_{\textnormal{q}j}} \mK_{ij} - \mK_{\textnormal{LCN}, i,j} = {}&\\
    e^{-(D-2r)/\lambda} (&1 - e^{-2r/\lambda} (2 - e^{-2r/\lambda})\\
    &+ \mathcal{O}(e^{-2D/\lambda})).
    \end{split}
\end{align}
\itshape
\end{theorem}

This shows that the error in $\mK_\textnormal{Nys}$ is close to $1$ for any reasonably large $\frac{r}{\lambda}$ (which is the maximum error possible). The errors in $\mK^\textnormal{sp}$ and $\mK_\textnormal{LCN}$ on the other hand are vanishingly small in this case, since $r \ll D$.\\
The reduced maximum error directly translates to an improved Sinkhorn approximation. We can show this by adapting the Sinkhorn approximation error bounds due to \citet{altschuler_massively_2019}.

\begin{definition}
\label{def:support}
A generalized diagonal is the set of elements $\mM_{i\sigma(i)} \ \forall i \in \{1,\ldots,n\}$ with matrix $\mM \in \R^{n \times n}$ and permutation $\sigma$. A non-negative matrix has support if it has a strictly positive generalized diagonal. It has total support if $\mM \neq 0$ and all non-zero elements lie on a strictly positive generalized diagonal.
\end{definition}

\begin{theorem}
\label{th:sinkhorn_error}
Let $X_\textnormal{p}, X_\textnormal{q} \subseteq \R^d$ have $n$ samples. Denote $\rho$ as the maximum distance between two samples. Let $\tilde{\mK}$ be a non-negative matrix with support, which approximates the similarity matrix $\mK$ with $\mK_{ij} = e^{-\|x_{\textnormal{p}i} - x_{\textnormal{q}j}\|_2/\lambda}$ and $\max_{i,j} |\tilde{\mK}_{ij} - \mK_{ij}| \le \frac{\varepsilon'}{2} e^{-\rho/\lambda}$, where $\varepsilon' = \min(1, \frac{\varepsilon}{50(\rho + \lambda \log \frac{\lambda n}{\varepsilon})})$. When performing the Sinkhorn algorithm until $\| \tilde{\mP} \vone_N - \vp\|_1 + \| \tilde{\mP}^T \vone_N - \vq \|_1 \le \varepsilon' / 2$, the resulting approximate transport plan $\tilde{\mP}$ and distance $\tilde{d}_c^\lambda$ are bounded by
\upshape
\begin{equation}
    | d_c^\lambda - \tilde{d}_{\tilde{c}}^\lambda | \le \varepsilon, \qquad
    \KL(\bar{\mP} \| \tilde{\mP}) \le \varepsilon / \lambda.
\end{equation}
\itshape
\end{theorem}

\textbf{Convergence rate.} We next show that sparse and LCN-Sinkhorn converge as fast as regular Sinkhorn by adapting the convergence bound by \citet{dvurechensky_computational_2018} to account for sparsity.

\begin{theorem}
\label{th:convergence}
Given a non-negative matrix $\tilde{\mK} \in \R^{n \times n}$ with support and $\vp \in \R^{n}$, $\vq \in \R^{n}$. The Sinkhorn algorithm gives a transport plan satisfying $\| \tilde{\mP} \vone_N - \vp \|_1 + \| \tilde{\mP}^T \vone_N - \vq \|_1 \le \varepsilon$ in iterations
\upshape
\begin{equation}
    k \le 2 + \frac{-4 \ln (\min_{i,j} \{ \tilde{\mK}_{ij} | \tilde{\mK}_{ij} > 0 \} \min_{i,j} \{ \vp_i, \vq_j \} )}{\varepsilon}.
\end{equation}
\itshape
\end{theorem}

\textbf{Backpropagation.} Efficient gradient computation is almost as important for modern deep learning models as the algorithm itself. These models usually aim at learning the embeddings in $X_\text{p}$ and $X_\text{q}$ and therefore need gradients w.r.t.\ the cost matrix $\mC$. We can estimate these either via automatic differentiation of the unrolled Sinkhorn iterations or via the analytic solution that assumes exact convergence. Depending on the problem at hand, either the automatic or the analytic estimator will lead to faster overall convergence \citep{ablin_super-efficiency_2020}. LCN-Sinkhorn works flawlessly with automatic backpropagation since it only relies on basic linear algebra (except for choosing Nyström landmarks and LSH neighbors, for which we use a simple straight-through estimator \citep{bengio_estimating_2013}). To enable fast analytic backpropagation we provide analytic gradients in \cref{prop:derivatives}. Note that both backpropagation methods have runtime linear in the number of points $n$ and $m$.

\begin{proposition}
\label{prop:derivatives}
In entropy-regularized OT and LCN-Sinkhorn the derivatives of the distances $d_c^\lambda$ and $d_{\textnormal{LCN},c}^\lambda$ (\cref{eq:entropy-ot,eq:lcn-sinkhorn}) and the optimal transport plan $\bar{\mP} \in \R^{n \times m}$ w.r.t.\ the (decomposed) cost matrix $\mC \in \R^{n \times m}$ with total support are
\upshape
\begin{gather}
    \frac{\partial d_c^\lambda}{\partial \mC} = \bar{\mP},\\
\begin{split}
    \frac{\partial d_{\text{LCN},c}^\lambda}{\partial \mU} = -\lambda \bar{\vs} (\mW \bar{\vt})^T,
    \quad& \frac{\partial d_{\text{LCN},c}^\lambda}{\partial \mW} = -\lambda (\bar{\vs}^T \mU)^T \bar{\vt}^T,\\
    \frac{\partial d_{\text{LCN},c}^\lambda}{\partial \log \mK^\text{sp}} = -\lambda \bar{\mP}^\text{sp},
    \quad& \frac{\partial d_{\text{LCN},c}^\lambda}{\partial \log \mK^\text{sp}_\text{Nys}} = -\lambda \bar{\mP}^\text{sp}_\text{Nys}.
\end{split}
\end{gather}
\itshape
This allows backpropagation in time $\mathcal{O}((n+m)l^2)$.
\end{proposition}

\section{Graph transport network} \label{sec:gtn}

\textbf{Graph distance learning.} Predicting similarities or distances between graph-structured objects is useful across a wide range of applications. It can be used to predict the reaction rate between molecules \citep{houston_machine_2019}, or search for similar images \citep{johnson_image_2015}, similar molecules for drug discovery \citep{birchall_training_2006}, or similar code for vulnerability detection \citep{li_graph_2019}. We propose the graph transport network (GTN) to evaluate approximate Sinkhorn on a full deep learning model and advance the state of the art on this task.

\textbf{Graph transport network.} GTN uses a Siamese graph neural network (GNN) to embed two graphs independently as \emph{sets} of node embeddings. These sets are then matched using multi-head unbalanced OT. Node embeddings represent the nodes' local environments, so similar neighborhoods will be close in embedding space and matched accordingly. Since Sinkhorn is symmetric and permutation invariant, any identical pair of graphs will thus by construction have a predicted distance of 0 (ignoring the entropy offset).\\
More precisely, given an undirected graph $\mathcal{G} = (\mathcal{V}, \mathcal{E})$, with node set $\mathcal{V}$ and edge set $\mathcal{E}$, node attributes $\vx_i \in \R^{H_\text{x}}$ and (optional) edge attributes $\ve_{i,j} \in \R^{H_\text{e}}$, with $i, j \in \mathcal{V}$, we update the node embeddings in each GNN layer via
\begin{align}
    \vh^{(l)}_{\text{self}, i} &= \sigma(\mW_\text{node}^{(l)} \vh^{(l-1)}_i + \vb^{(l)}),\\
    \vh^{(l)}_i &= \vh^{(l)}_{\text{self}, i} + \sum_{j \in \mathcal{N}_i} \eta^{(l)}_{i, j} \vh^{(l)}_{\text{self}, j} \tW_\text{edge} \ve_{i,j},
\end{align}
with $\mathcal{N}_i$ denoting the neighborhood of node $i$, $\vh^{(0)}_i = \vx_i$, $\vh^{(l)}_i \in \R^{H_\text{N}}$ for $l \ge 1$, the bilinear layer $\tW_\text{edge} \in \R^{H_\text{N} \times H_\text{N} \times H_\text{e}}$, and the degree normalization $\eta^{(1)}_{i, j} = 1$ and $\eta^{(l)}_{i, j} = 1 / \sqrt{\operatorname{deg}_i \operatorname{deg}_j}$ for $l > 1$. This choice of $\eta_{i, j}$ allows our model to handle highly skewed degree distributions while still being able to represent node degrees. We found the choice of non-linearity $\sigma$ not to be critical and chose a LeakyReLU. We do not use the bilinear layer $\tW_\text{edge} \ve_{i,j}$ if there are no edge attributes. We aggregate each layer's node embeddings to obtain the overall embedding of node $i$
\begin{equation}
    \vh_i^\text{GNN} = [ \vh^{(1)}_{\text{self}, i} \,\|\, \vh^{(1)}_i \,\|\, \vh^{(2)}_i \,\|\, \dots \,\|\, \vh^{(L)}_i ].
\end{equation}
We then compute the embeddings for matching via $\vh_i^\text{final} = \operatorname{MLP}(\vh_i^\text{GNN})$. Having obtained the embedding sets $H_1^\text{final}$ and $H_2^\text{final}$ of both graphs we use the $L_2$ distance as a cost function for the Sinkhorn distance. Finally, we calculate the prediction from the Sinkhorn distance via $d = d_c^\lambda w_\text{out} + b_\text{out}$, with learnable $w_\text{out}$ and $b_\text{out}$. GTN is trained end-to-end via backpropagation. For small graphs we use the full Sinkhorn distance and scale to large graphs with LCN-Sinkhorn. GTN is more expressive than models that aggegrate node embeddings to a single fixed-size embedding but still scales log-linearly in the number of nodes, as opposed to previous approaches which scale quadratically.

\textbf{Learnable unbalanced OT.} Since GTN regularly encounters graphs with disagreeing numbers of nodes it needs to be able to handle cases where $\|\vp\|_1 \neq \|\vq\|_1$ or where not all nodes in one graph have a corresponding node in the other, i.e.\ $\mP \vone_m < \vp$ or $\mP^T \vone_n < \vq$. Unbalanced OT allows handling both of these cases \citep{peyre_computational_2019}, usually by swapping the strict balancing requirements with a uniform divergence loss term on $\vp$ and $\vq$ \citep{frogner_learning_2015,chizat_scaling_2018}. However, this \emph{uniformly} penalizes deviations from balanced OT and therefore cannot adaptively ignore parts of the distribution. We propose to improve on this by swapping the cost matrix $\mC$ with the bipartite matching (BP) matrix \citep{riesen_approximate_2009}
\begin{equation}
\begin{split}
    \mC_\text{BP} = \begin{bmatrix}
        \mC & \mC^{(\text{p},\varepsilon)} \\
        \mC^{(\varepsilon,\text{q})} & \mC^{(\varepsilon,\varepsilon)}
        \end{bmatrix}, \quad&
    \mC^{(\text{p},\varepsilon)}_{ij} = \begin{cases}
        c_{i, \varepsilon} & i = j\\
        \infty & i \neq j
    \end{cases},\\
    \mC^{(\varepsilon,\text{q})}_{ij} = \begin{cases}
        c_{\varepsilon, j} & i = j\\
        \infty & i \neq j
    \end{cases}, \quad&
    \mC^{(\varepsilon,\varepsilon)}_{ij} = 0,
\end{split}
\end{equation}
and obtain the deletion cost $c_{i, \varepsilon}$ and $c_{\varepsilon, j}$ from the input sets $X_\text{p}$ and $X_\text{q}$. Using the BP matrix only adds minor computational overhead since we just need to save the diagonals $\vc_{\text{p},\varepsilon}$ and $\vc_{\varepsilon,\text{q}}$ of $\mC_{\text{p},\varepsilon}$ and $\mC_{\varepsilon,\text{q}}$. We can then use $\mC_\text{BP}$ in the Sinkhorn algorithm (\cref{eq:sinkhorn}) via
\begin{equation}
    \mK_\text{BP} \vt = \begin{bmatrix}
        \mK \hat{\vt} + \vc_{\text{p},\varepsilon} \odot \check{\vt} \\
        \vc_{\varepsilon,\text{q}} \odot \hat{\vt} + \vone_n^T \check{\vt}
    \end{bmatrix}, \
    \mK_\text{BP}^T \vs = \begin{bmatrix}
        \mK^T \hat{\vs} + \vc_{\varepsilon,\text{q}} \odot \check{\vs} \\
        \vc_{\text{p},\varepsilon} \odot \hat{\vs} + \vone_m^T \check{\vs}
    \end{bmatrix},
\end{equation}
where $\hat{\vt}$ denotes the upper and $\check{\vt}$ the lower part of the vector $\vt$. To calculate $d_c^\lambda$ we can decompose the transport plan $\mP_\text{BP}$ in the same way as $\mC_\text{BP}$, with a single scalar for $\mP_{\varepsilon,\varepsilon}$. For GTN we obtain the deletion cost via $c_{i,\varepsilon} = \| \valpha \odot \vx_{\text{p}i} \|_2$, with a learnable vector $\valpha \in \R^d$.

\textbf{Multi-head OT.} Inspired by attention models \citep{vaswani_attention_2017} and multiscale kernels \citep{bermanis_multiscale_2013} we further improve GTN by using multiple OT heads. Using $K$ heads means that we calculate $K$ separate sets of embeddings representing the same pair of objects by using separate linear layers, i.e.\ $\vh_{k, i}^{\text{final}} = \mW^{(k)} \vh_i^\text{GNN}$ for head $k$. We then calculate OT in parallel for these sets using a series of regularization parameters $\lambda_k = 2^{k - K/2} \lambda$. This yields a set of distances $\vd_c^\lambda \in \R^K$. We obtain the final prediction via $d = \operatorname{MLP}(\vd_c^\lambda)$. Both learnable unbalanced OT and multi-head OT might be of independent interest.


\section{Related work}

\textbf{Hierarchical kernel approximation.} These methods usually hierarchically decompose the kernel matrix into separate blocks and use low-rank or core-diagonal approximations for each block \citep{si_memory_2017,ding_multiresolution_2017}. This idea is similar in spirit to LCN, but LCN boils it down to its essence by using one purely global part and a fine-grained LSH method to obtain one exact and purely local part.

\textbf{Log-linear optimal transport.} For an overview of optimal transport and its foundations see \citet{peyre_computational_2019}. On low-dimensional grids and surfaces OT can be solved using dynamical OT \citep{papadakis_optimal_2014,solomon_earth_2014}, convolutions \citep{solomon_convolutional_2015}, or embedding/hashing schemes \citep{indyk_fast_2003,andoni_earth_2008}. In higher dimensions we can use tree-based algorithms \citep{backurs_scalable_2020} or hashing schemes \citep{charikar_similarity_2002}, which are however limited to a previously fixed set of points $X_\text{p}$, $X_\text{q}$, on which only the distributions $\vp$ and $\vq$ change. Another approach are sliced Wasserstein distances \citep{rabin_wasserstein_2011}. However, they do not provide a transport plan, require the $L_2$ distance as a cost function, and are either unstable in convergence or prohibitively expensive for high dimensions ($\mathcal{O}(nd^3)$) \citep{meng_large-scale_2019}. For high-dimensional sets that change dynamically (e.g.\ during training) one method of achieving log-linear runtime is a multiscale approximation of entropy-regularized OT \citep{schmitzer_stabilized_2019,gerber_multiscale_2017}. \citet{tenetov_fast_2018} recently proposed using a low-rank approximation of the Sinkhorn similarity matrix obtained via a semidiscrete approximation of the Euclidean distance. \citet{altschuler_massively_2019} improved upon this approach by using the Nyström method for the approximation. However, these approaches still struggle with high-dimensional real-world problems, as we will show in \cref{sec:exp}.

\textbf{Accelerating Sinkhorn.} Another line of work has been pursuing accelerating entropy-regularized OT without changing its computational complexity w.r.t.\ the number of points. Original Sinkhorn requires $\mathcal{O}(1/\varepsilon^2)$ iterations, but \citet{dvurechensky_computational_2018} and \citet{jambulapati_direct_2019} recently proposed algorithms that reduce the computational complexity to $\mathcal{O}(\min(n^{9/4}/\varepsilon, n^2/\varepsilon^2))$ and $\mathcal{O}(n^2/\varepsilon)$, respectively. \citet{mensch_online_2020} proposed an online Sinkhorn algorithm to significantly reduce its memory cost. \citet{alaya_screening_2019} proposed reducing the size of the Sinkhorn problem by screening out neglectable components, which allows for approximation guarantees. \citet{genevay_stochastic_2016} proposed using a stochastic optimization scheme instead of Sinkhorn iterations. \citet{essid_quadratically_2018} and \citet{blondel_smooth_2018} proposed alternative regularizations to obtain OT problems with similar runtimes as the Sinkhorn algorithm. This work is largely orthogonal to ours.

\textbf{Embedding alignment.} For an overview of cross-lingual word embedding models see \citet{ruder_survey_2019}. Unsupervised word embedding alignment was proposed by \citet{conneau_word_2018}, with subsequent advances by \citet{alvarez-melis_gromov-wasserstein_2018,grave_unsupervised_2019,joulin_loss_2018}.

\textbf{Graph matching and distance learning.} Graph neural networks (GNNs) have recently been successful on a wide variety of graph-based tasks \citep{kipf_semi-supervised_2017,gasteiger_predict_2019,gasteiger_directional_2020,zambaldi_deep_2019}. GNN-based approaches for graph matching and graph distance learning either rely on a single fixed-dimensional graph embedding \citep{bai_simgnn_2019,li_graph_2019}, or only use attention or some other strongly simplified variant of optimal transport \citep{bai_simgnn_2019,riba_learning_2018,li_graph_2019}. Others break permutation invariance and are thus ill-suited for this task \citep{ktena_distance_2017,bai_convolutional_2018}. So far only approaches using a single graph embedding allow faster than quadratic scaling in the number of nodes. Compared to the Sinkhorn-based image model proposed by \citet{wang_learning_2019} GTN uses no CNN or cross-graph attention, but an enhanced GNN and embedding aggregation scheme. OT has recently been proposed for graph kernels \citep{maretic_got_2019,vayer_optimal_2019}, which can (to some extent) be used for graph matching, but not for distance learning.

\section{Experiments} \label{sec:exp}

\begin{table*}[t]
    \centering
    \sisetup{table-format=\pm1.3}
    \caption{Mean and standard deviation of relative Sinkhorn distance error, IoU of top \SI{0.1}{\percent} and correlation coefficient (PCC) of OT plan entries across 5 runs. Sparse Sinkhorn and LCN-Sinkhorn achieve the best approximation in all 3 measures.}
    \resizebox{\textwidth}{!}{
    \begin{tabular}{=l+S+S@{\hspace{0.2cm}}+S+S+S+S@{\hspace{0.2cm}}+S+S+S+S@{\hspace{0.2cm}}+S+S+S+S@{\hspace{0.2cm}}+S}
{} & \multicolumn{3}{c}{EN-DE} && \multicolumn{3}{c}{EN-ES} && \multicolumn{3}{c}{3D point cloud} && \multicolumn{3}{c}{Uniform in $d$-ball ($d\!=\!16$)} \\
\cline{2-4} \cline{6-8} \cline{10-12} \cline{14-16}
{} &        \mcc{Rel. err. $d_c^\lambda$} &                            \mcc{PCC} &                            \mcc{IoU} &              &        \mcc{Rel. err. $d_c^\lambda$} &                            \mcc{PCC} &                            \mcc{IoU} &              &        \mcc{Rel. err. $d_c^\lambda$} &                            \mcc{PCC} &                            \mcc{IoU} &               &        \mcc{Rel. err. $d_c^\lambda$} &                            \mcc{PCC} &                            \mcc{IoU} \\
\midrule
Factored OT   & 0.318           & 0.044           & 0.019           &  & 0.332           & 0.037           & 0.026           &  & 6.309         & 0.352           & 0.004           &  & 1.796         & 0.096           & 0.029           \\
\vspace{2pt} \rowstyle{\color{gray}}    & +- 0.001          & +- 0.001          & +- 0.002          &  & +- 0.001          & +- 0.002          & +- 0.005          &  & +- 0.004        & +- 0.001          & +- 0.001          &  & +- 0.001        & +- 0.001          & +- 0.000          \\

Multiscale OT & 0.634           & 0.308           & 0.123           &  & 0.645           & 0.321           & 0.125           &  & \bfseries 0.24  & 0.427           & 0.172           &  & \bfseries 0.03  & 0.091           & 0.021           \\
\vspace{2pt} \rowstyle{\color{gray}} & +- 0.011          & +- 0.014          & +- 0.005          &  & +- 0.014          & +- 0.006          & +- 0.012          &  & \bfseries +- 0.07 & +- 0.008          & +- 0.011          &  & \bfseries +- 0.02 & +- 0.005          & +- 0.001          \\

Nyström Skh.  & 1.183           & 0.077           & 0.045           &  & 1.175           & 0.068           & 0.048           &  & 1.89         & \bfseries 0.559  & 0.126           &  & 1.837         & 0.073           & 0.018           \\
\vspace{2pt} \rowstyle{\color{gray}}  & +- 0.005          & +- 0.001          & +- 0.005          &  & +- 0.018          & +- 0.001          & +- 0.006          &  & +- 0.07        & \bfseries +- 0.009 & +- 0.014          &  & +- 0.006        & +- 0.000          & +- 0.000          \\

Sparse Skh.   & \bfseries 0.233  & 0.552           & 0.102           &  & \bfseries 0.217  & 0.623           & 0.102           &  & 0.593         & 0.44             & 0.187           &  & 0.241         & \bfseries 0.341  & \bfseries 0.090  \\
\vspace{2pt} \rowstyle{\color{gray}}   & \bfseries +- 0.002 & +- 0.004          & +- 0.001          &  & \bfseries +- 0.001 & +- 0.004          & +- 0.001          &  & +- 0.015        & +- 0.03            & +- 0.014          &  & +- 0.002        & \bfseries +- 0.004 & \bfseries +- 0.001 \\

LCN-Sinkhorn  & 0.406           & \bfseries 0.673  & \bfseries 0.197  &  & 0.368           & \bfseries 0.736  & \bfseries 0.201  &  & 1.91         & \bfseries 0.564  & \bfseries 0.195  &  & 0.435         & 0.328           & 0.079    \\
\rowstyle{\color{gray}}  & +- 0.015          & \bfseries +- 0.012 & \bfseries +- 0.007 &  & +- 0.012          & \bfseries +- 0.003 & \bfseries +- 0.003 &  & +- 0.28        & \bfseries +- 0.008 & \bfseries +- 0.013 &  & +- 0.009        & +- 0.006          & +- 0.001   
\end{tabular}
    }
    \label{tab:sinkhorn}
\end{table*}

\begin{figure*}[t]
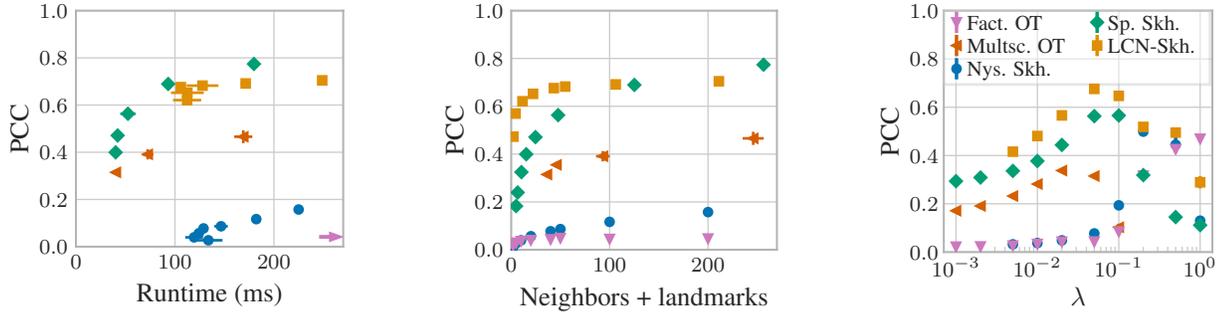

    \centering
    \begin{subfigure}[t]{0.325\textwidth}
        \input{figures/pcc_runtime.pgf}
    \end{subfigure}
    \hfill
    \begin{subfigure}[t]{0.325\textwidth}
        \input{figures/pcc_neighbors.pgf}
    \end{subfigure}
    \hfill
    \begin{subfigure}[t]{0.325\textwidth}
        \input{figures/pcc_reg.pgf}
    \end{subfigure}
    \caption{OT plan approximation quality for EN-DE, via PCC. \textbf{Left:} Sparse Sinkhorn offers the best tradeoff with runtime, with LCN-Sinkhorn closely behind. \textbf{Center:} LCN-Sinkhorn achieves the best approximation for low and sparse Sinkhorn for high numbers of neighbors/landmarks. \textbf{Right:} Sparse Sinkhorn performs best for low, LCN-Sinkhorn for moderate and factored OT for very high entropy regularization $\lambda$. The arrow indicates factored OT results far outside the range.}
    \label{fig:pcc}
\end{figure*}

\textbf{Approximating Sinkhorn.} We start by directly investigating different Sinkhorn approximations. To do so we compute entropy-regularized OT on (i) pairs of $10^4$ word embeddings from \citet{conneau_word_2018}, which we preprocess with Wasserstein Procrustes alignment in order to obtain both close and distant neighbors, (ii) the armadillo and dragon point clouds from the Stanford 3D Scanning Repository \citep{stanford_stanford_2014} (with $10^4$ randomly subsampled points), and (iii) pairs of $10^4$ data points that are uniformly distributed in the $d$-ball ($d=16$). We let every method use the same total number of 40 average neighbors and landmarks (LCN uses 20 each) and set $\lambda=0.05$ (as e.g.\ in \citet{grave_unsupervised_2019}). Besides the Sinkhorn distance we measure transport plan approximation quality by (a) calculating the Pearson correlation coefficient (PCC) between all entries in the approximated plan and the true $\bar{\mP}$ and (b) comparing the sets of \SI{0.1}{\percent} largest entries in the approximated and true $\bar{\mP}$ using the Jaccard similarity (intersection over union, IoU). Note that usually the OT plan is more important than the distance, since it determines the training gradient and tasks like embedding alignment are exclusively based on the OT plan. In all figures the error bars denote standard deviation across 5 runs, which is often too small to be visible.

\cref{tab:sinkhorn} shows that for word embeddings both sparse Sinkhorn, LCN-Sinkhorn and factored OT \citep{forrow_statistical_2019} obtain distances that are significantly closer to the true $d_c^\lambda$ than Multiscale OT and Nyström-Sinkhorn. Furthermore, the transport plan computed by sparse Sinkhorn and LCN-Sinkhorn show both a PCC and IoU that are around twice as high as Multiscale OT, while Nyström-Sinkhorn and factored OT exhibit almost no correlation. LCN-Sinkhorn performs especially well in this regard. This is also evident in \cref{fig:ot-plan}, which shows how the $10^4 \times 10^4$ approximated OT plan entries compared to the true Sinkhorn values. Multiscale OT shows the best distance approximation on 3D point clouds and random high-dimensional data. However, sparse Sinkhorn and LCN-Sinkhorn remain the best OT plan approximations, especially in high dimensions.

\cref{fig:pcc} shows that sparse Sinkhorn offers the best trade-off between runtime and OT plan quality. Factored OT exhibits a runtime \numrange{2}{10} times longer than the competition due to its iterative refinement scheme. LCN-Sinkhorn performs best for use cases with constrained memory (few neighbors/landmarks). The number of neighbors and landmarks directly determines memory usage and is linearly proportional to the runtime (see \cref{app:runtimes}). \cref{fig:pcc} furthermore shows that sparse Sinkhorn performs best for low regularizations, where LCN-Sinkhorn fails due to the Nyström part going out of bounds. Nyström Sinkhorn performs best at high values and LCN-Sinkhorn always performs better than both (as long as it can be calculated). Interestingly, all approximations except factored OT seem to fail at high $\lambda$. We defer analogously discussing the distance approximation to \cref{app:distance}. All approximations scale linearly both in the number of neighbors/landmarks and dataset size, as shown in \cref{app:runtimes}. Overall, we see that sparse Sinkhorn and LCN-Sinkhorn yield significant improvements over previous approximations. However, do these improvements also translate to better performance on downstream tasks?

\begin{table*}[t]
    \centering
    \sisetup{table-format=2.1(2)}
    \caption{Accuracy and standard deviation across 5 runs for unsupervised word embedding alignment with Wasserstein Procrustes. LCN-Sinkhorn improves upon the original by 3.1 pp. before and 2.0 pp. after iterative CSLS refinement. {\scriptsize *Migrated and re-run on GPU via PyTorch}}
    \resizebox{\textwidth}{!}{
    \begin{tabular}{lS[table-format=3.1]SSSSSSSSS[table-format=2.1]}
{} &       \mcc{Time (s}) &                        \mcc{EN-ES} &                        \mcc{ES-EN} &                        \mcc{EN-FR} &                        \mcc{FR-EN} &                        \mcc{EN-DE} &                        \mcc{DE-EN} &                        \mcc{EN-RU} &                        \mcc{RU-EN} &                 \mcc{\textbf{Avg.}} \\
\hline
Original*                    &            268 &           79.2 \pm 0.2 &           78.8 \pm 2.8 &           81.0 \pm 0.3 &           79.4 \pm 0.9 &           71.7 \pm 0.2 &           65.7 \pm 3.4 &           36.3 \pm 1.1 &           51.1 \pm 1.1 &           67.9 \\
Full Sinkhorn                &            402 &           81.1 \pm 0.0 &  \bfseries 82.0 \pm 0.0 &           81.2 \pm 0.0 &           81.3 \pm 0.0 &  \bfseries 74.1 \pm 0.0 &           70.7 \pm 0.0 &           37.3 \pm 0.0 &           53.5 \pm 0.0 &           70.1 \\
Multiscale OT                &     88.2 &          24 \pm 31 &          74.7 \pm 3.3 &          27 \pm 32 &            6.3 \pm 4.4 &          36 \pm 10 &          47 \pm 21 &            0.0 \pm 0.0 &            0.2 \pm 0.1 &           26.8 \\
Nyström Skh.                 &            102 &           64.4 \pm 1.0 &           59.3 \pm 1.2 &           64.1 \pm 1.6 &           56.8 \pm 4.0 &           54.1 \pm 0.6 &           47.1 \pm 3.5 &           14.1 \pm 1.2 &           22.5 \pm 2.4 &           47.8 \\
\textbf{Sparse Skh.}         &     49.2 &           80.2 \pm 0.2 &           81.7 \pm 0.4 &           80.9 \pm 0.3 &           80.1 \pm 0.2 &           72.1 \pm 0.6 &           65.1 \pm 1.7 &           35.5 \pm 0.6 &           51.5 \pm 0.4 &           68.4 \\
\textbf{LCN-Sinkhorn}       &     86.8 &  \bfseries 81.8 \pm 0.2 &  \bfseries 81.3 \pm 1.8 &  \bfseries 82.0 \pm 0.4 &  \bfseries 82.1 \pm 0.3 &           73.6 \pm 0.2 &  \bfseries 71.3 \pm 0.9 &  \bfseries 41.0 \pm 0.8 &  \bfseries 55.1 \pm 1.4 &  \bfseries 71.0 \\
\hline
Original* + ref.             &   \mcc{268+81} &           83.0 \pm 0.3 &  \bfseries 82.0 \pm 2.5 &  \bfseries 83.8 \pm 0.1 &           83.0 \pm 0.4 &  \bfseries 77.3 \pm 0.3 &           69.7 \pm 4.3 &           46.2 \pm 1.0 &           54.0 \pm 1.1 &           72.4 \\
\bfseries LCN-Skh. + ref. &  \mcc{86.8+81} &  \bfseries 83.5 \pm 0.2 &  \bfseries 83.1 \pm 1.3 &  \bfseries 83.8 \pm 0.2 &  \bfseries 83.6 \pm 0.1 &  \bfseries 77.2 \pm 0.3 &  \bfseries 72.8 \pm 0.7 &  \bfseries 51.8 \pm 2.6 &  \bfseries 59.2 \pm 1.9 &  \bfseries 74.4 \\
\end{tabular}
    }
    \label{tab:emb}
\end{table*}

\textbf{Embedding alignment.} Embedding alignment is the task of finding the orthogonal matrix $\mR \in \R^{d \times d}$ that best aligns the vectors from two different embedding spaces, which is e.g.\ useful for unsupervised word translation. We use the experimental setup established by \citet{conneau_word_2018} by migrating \citet{grave_unsupervised_2019}'s implementation to PyTorch. The only change we make is using the full set of \num{20000} word embeddings and training for 300 steps, while reducing the learning rate by half every 100 steps. We do not change \emph{any} other hyperparameters and do not use unbalanced OT. After training we match pairs via cross-domain similarity local scaling (CSLS) \citep{conneau_word_2018}. We use 10 Sinkhorn iterations, 40 neighbors on average for sparse Sinkhorn, and 20 neighbors and landmarks for LCN-Sinkhorn (for details see \cref{app:landmarks}). We allow both multiscale OT and Nyström Sinkhorn to use as many landmarks and neighbors as can fit into GPU memory and finetune both methods.

\cref{tab:emb} shows that using full Sinkhorn yields a significant improvement in accuracy on this task compared to the original approach of performing Sinkhorn on randomly sampled subsets of embeddings \citep{grave_unsupervised_2019}. LCN-Sinkhorn even outperforms the \emph{full} version in most cases, which is likely due to regularization effects from the approximation. It also runs 4.6x faster than full Sinkhorn and 3.1x faster than the original scheme, while using \SI{88}{\percent} and \SI{44}{\percent} less memory, respectively. Sparse Sinkhorn runs 1.8x faster than LCN-Sinkhorn but cannot match its accuracy. LCN-Sinkhorn still outcompetes the original method after refining the embeddings with iterative local CSLS \citep{conneau_word_2018}. Both multiscale OT and Nyström Sinkhorn fail at this task, despite their larger computational budget. This shows that the improvements achieved by sparse Sinkhorn and LCN-Sinkhorn have an even larger impact in practice.

\textbf{Graph distance regression.} The graph edit distance (GED) is useful for various tasks such as image retrieval \citep{xiao_hmm-based_2008} or fingerprint matching \citep{neuhaus_error-tolerant_2004}, but its computation is NP-complete \citep{bunke_graph_1998}.
For large graphs we therefore need an effective approximation. We use the Linux dataset by \citet{bai_simgnn_2019} and generate 2 new datasets by computing the exact GED using the method by \citet{lerouge_new_2017} on small graphs ($\leq 30$ nodes) from the AIDS dataset \citep{riesen_iam_2008} and a set of preferential attachment graphs. We compare GTN to 3 state-of-the-art baselines: SiameseMPNN \citep{riba_learning_2018}, SimGNN \citep{bai_simgnn_2019}, and the Graph Matching Network (GMN) \citep{li_graph_2019}. We tune the hyperparameters of all baselines and GTN via grid search. For more details see \cref{app:landmarks,app:implementation,app:graph}.

We first test GTN and the proposed OT enhancements. \cref{tab:graph-distance} shows that GTN improves upon other models by \SI{20}{\percent} with a single head and by \SI{48}{\percent} with 8 OT heads. Its performance breaks down with regular unbalanced (using KL-divergence loss for the marginals) and balanced OT, showing the importance of learnable unbalanced OT.

\begin{table}[t]
    \centering
    \caption{RMSE for GED regression across 3 runs and the targets' standard deviation $\sigma$. GTN outperforms previous models by \SI{48}{\percent}.}
    \begin{tabular}{lS[table-format=1.3(1)]S[table-format=2.1(1)]S[table-format=2.1(1)]}
              & \mcc{Linux} & \mcc{AIDS30}                    & \mcc{Pref.\ att.}                \\ \hline
$\sigma$      & 0.184      & 16.2                & 48.3                \\
SiamMPNN      & 0.090 +- 0.007 & 13.8 +- 0.3         & 12.1 +- 0.6         \\
SimGNN        & 0.039 & 4.5 +- 0.3          & 8.3 +- 1.4          \\
GMN           & 0.015 +- 0.000 & 10.3 +- 0.6         & 7.8 +- 0.3          \\ \hline
GTN, 1 head   & 0.022 +- 0.001 & 3.7 +- 0.1          & 4.5 +- 0.3          \\
8 OT heads    & \bfseries 0.012 +- 0.001 & \bfseries 3.2 +- 0.1 & \bfseries 3.5 +- 0.2 \\
Unbalanced OT   & 0.033 +- 0.002 & 15.7 +- 0.5         & 9.7 +- 0.9 \\
Balanced OT   & 0.034 +- 0.001 & 15.3 +- 0.1         & 27.4 +- 0.9
\end{tabular}
    \label{tab:graph-distance}
\end{table}

\begin{table}[t]
    \centering
    \sisetup{table-format=2.1(2)}
    \caption{RMSE for graph distance regression across 3 runs and the targets' standard deviation $\sigma$. Using LCN-Sinkhorn with GTN increases the error by only \SI{10}{\percent} and allows log-linear scaling.}
    \begin{tabular}{lSSS[table-format=2.2(1)]}
              & \multicolumn{2}{c}{GED}                               & \mcc{PM [$10^{-2}$]}              \\ \cline{2-3}
              & \mcc{AIDS30}                    & \mcc{Pref.\ att.}                & \mcc{Pref.\ att.\ 200}              \\ \hline
$\sigma$      & 16.2                      & 48.3                      & 10.2                        \\
Full Sinkhorn & 3.7 +- 0.1          & 4.5 +- 0.3          & 1.27 +- 0.06          \\
\hline
Nyström Skh.  & \bfseries 3.6 +- 0.3          & 6.2 +- 0.6          & 2.43 +- 0.07          \\
Multiscale OT & 11.2 +- 0.3         & 27.4 +- 5.4               & 6.71 +- 0.44          \\
Sparse Skh.   & 44 +- 30              & 40.7 +- 8.1              & 7.57 +- 1.09          \\
LCN-Skh.      & 4.0 +- 0.1          & \bfseries 5.1 +- 0.4          & \bfseries 1.41 +- 0.15
\end{tabular}
    \label{tab:gtn-approx}
\end{table}

Having established GTN as a state-of-the-art model we next ask whether we can sustain its performance when using approximate OT. For this we additionally generate a set of larger graphs with around 200 nodes and use the Pyramid matching (PM) kernel \citep{nikolentzos_matching_2017} as the prediction target, since these graphs are too large to compute the GED. See \cref{app:graph} for hyperparameter details. \cref{tab:gtn-approx} shows that both sparse Sinkhorn and the multiscale method using 4 (expected) neighbors fail at this task, demonstrating that the low-rank approximation in LCN has a crucial stabilizing effect during training. Nyström Sinkhorn with 4 landmarks performs surprisingly well on the AIDS30 dataset, suggesting an overall low-rank structure with Nyström acting as regularization. However, it does not perform as well on the other two datasets. Using LCN-Sinkhorn with 2 neighbors and landmarks works well on all three datasets, with an RMSE increased by only \SI{10}{\percent} compared to full GTN. \cref{app:runtimes} furthermore shows that GTN with LCN-Sinkhorn indeed scales linearly in the number of nodes across multiple orders of magnitude. This model thus enables graph matching and distance learning on graphs that are considered large even for simple node-level tasks (\num{20000} nodes).

\section{Conclusion}

Locality-sensitive hashing (LSH) and the novel locally corrected Nyström (LCN) method enable fast and accurate approximations of entropy-regularized OT with log-linear runtime: Sparse Sinkhorn and LCN-Sinkhorn. The graph transport network (GTN) is one example for such a model, which can be substantially improved with learnable unbalanced OT and multi-head OT. It sets the new state of the art for graph distance learning while still scaling log-linearly with graph size. These contributions enable new applications and models that are both faster and more accurate, since they can sidestep workarounds such as pooling.

\subsubsection*{Acknowledgments}
We would like to thank Johannes Pitz, Oleksandr Shchur, Aleksandar Bojchevski, and Daniel Zügner for their support, suggestions, and feedback during the process of creating this paper.
This research was supported by the Deutsche Forschungsgemeinschaft (DFG) through the Emmy Noether grant GU 1409/2-1 and the TUM International Graduate School of Science and Engineering (IGSSE), GSC 81.

\FloatBarrier
\bibliography{fast-ot}

\begin{thebibliography}{79}
\providecommand{\natexlab}[1]{#1}
\providecommand{\url}[1]{\texttt{#1}}
\expandafter\ifx\csname urlstyle\endcsname\relax
  \providecommand{\doi}[1]{doi: #1}\else
  \providecommand{\doi}{doi: \begingroup \urlstyle{rm}\Url}\fi

\bibitem[Ablin et~al.(2020)Ablin, Peyré, and
  Moreau]{ablin_super-efficiency_2020}
Pierre Ablin, Gabriel Peyré, and Thomas Moreau.
\newblock Super-efficiency of automatic differentiation for functions defined
  as a minimum.
\newblock In \emph{{ICML}}, 2020.

\bibitem[Alaya et~al.(2019)Alaya, Berar, Gasso, and
  Rakotomamonjy]{alaya_screening_2019}
Mokhtar~Z. Alaya, Maxime Berar, Gilles Gasso, and Alain Rakotomamonjy.
\newblock Screening {Sinkhorn} {Algorithm} for {Regularized} {Optimal}
  {Transport}.
\newblock In \emph{{NeurIPS}}, 2019.

\bibitem[Altschuler et~al.(2019)Altschuler, Bach, Rudi, and
  Niles-Weed]{altschuler_massively_2019}
Jason Altschuler, Francis Bach, Alessandro Rudi, and Jonathan Niles-Weed.
\newblock Massively scalable {Sinkhorn} distances via the {Nyström} method.
\newblock In \emph{{NeurIPS}}, 2019.

\bibitem[Alvarez-Melis \& Jaakkola(2018)Alvarez-Melis and
  Jaakkola]{alvarez-melis_gromov-wasserstein_2018}
David Alvarez-Melis and Tommi~S. Jaakkola.
\newblock Gromov-{Wasserstein} {Alignment} of {Word} {Embedding} {Spaces}.
\newblock In \emph{{EMNLP}}, 2018.

\bibitem[Andoni et~al.(2008)Andoni, Indyk, and Krauthgamer]{andoni_earth_2008}
Alexandr Andoni, Piotr Indyk, and Robert Krauthgamer.
\newblock Earth mover distance over high-dimensional spaces.
\newblock In \emph{{ACM}-{SIAM} symposium on {Discrete} algorithms ({SODA})},
  2008.

\bibitem[Andoni et~al.(2015)Andoni, Indyk, Laarhoven, Razenshteyn, and
  Schmidt]{andoni_practical_2015}
Alexandr Andoni, Piotr Indyk, Thijs Laarhoven, Ilya~P. Razenshteyn, and Ludwig
  Schmidt.
\newblock Practical and {Optimal} {LSH} for {Angular} {Distance}.
\newblock In \emph{{NeurIPS}}, 2015.

\bibitem[Arjovsky et~al.(2017)Arjovsky, Chintala, and
  Bottou]{arjovsky_wasserstein_2017}
Martin Arjovsky, Soumith Chintala, and Léon Bottou.
\newblock Wasserstein {Generative} {Adversarial} {Networks}.
\newblock In \emph{{ICML}}, 2017.

\bibitem[Backurs et~al.(2020)Backurs, Dong, Indyk, Razenshteyn, and
  Wagner]{backurs_scalable_2020}
Arturs Backurs, Yihe Dong, Piotr Indyk, Ilya Razenshteyn, and Tal Wagner.
\newblock Scalable {Nearest} {Neighbor} {Search} for {Optimal} {Transport}.
\newblock In \emph{{ICML}}, 2020.

\bibitem[Bai et~al.(2018)Bai, Ding, Sun, and Wang]{bai_convolutional_2018}
Yunsheng Bai, Hao Ding, Yizhou Sun, and Wei Wang.
\newblock Convolutional {Set} {Matching} for {Graph} {Similarity}.
\newblock In \emph{Relational {Representation} {Learning} {Workshop},
  {NeurIPS}}, 2018.

\bibitem[Bai et~al.(2019)Bai, Ding, Bian, Chen, Sun, and Wang]{bai_simgnn_2019}
Yunsheng Bai, Hao Ding, Song Bian, Ting Chen, Yizhou Sun, and Wei Wang.
\newblock {SimGNN}: {A} {Neural} {Network} {Approach} to {Fast} {Graph}
  {Similarity} {Computation}.
\newblock In \emph{{WSDM}}, 2019.

\bibitem[Bengio et~al.(2013)Bengio, Léonard, and
  Courville]{bengio_estimating_2013}
Yoshua Bengio, Nicholas Léonard, and Aaron Courville.
\newblock Estimating or {Propagating} {Gradients} {Through} {Stochastic}
  {Neurons} for {Conditional} {Computation}.
\newblock \emph{arXiv}, 1308.3432, 2013.

\bibitem[Berg et~al.(1984)Berg, Christensen, and Ressel]{berg_harmonic_1984}
Christian Berg, Jens Peter~Reus Christensen, and Paul Ressel.
\newblock \emph{Harmonic {Analysis} on {Semigroups}}.
\newblock Number 100 in Graduate {Texts} in {Mathematics}. 1984.

\bibitem[Bermanis et~al.(2013)Bermanis, Averbuch, and
  Coifman]{bermanis_multiscale_2013}
Amit Bermanis, Amir Averbuch, and Ronald~R. Coifman.
\newblock Multiscale data sampling and function extension.
\newblock \emph{Applied and Computational Harmonic Analysis}, 34\penalty0
  (1):\penalty0 15--29, 2013.

\bibitem[Birchall et~al.(2006)Birchall, Gillet, Harper, and
  Pickett]{birchall_training_2006}
Kristian Birchall, Valerie~J. Gillet, Gavin Harper, and Stephen~D. Pickett.
\newblock Training {Similarity} {Measures} for {Specific} {Activities}:
  {Application} to {Reduced} {Graphs}.
\newblock \emph{Journal of Chemical Information and Modeling}, 46\penalty0
  (2):\penalty0 577--586, 2006.

\bibitem[Blondel et~al.(2018)Blondel, Seguy, and Rolet]{blondel_smooth_2018}
Mathieu Blondel, Vivien Seguy, and Antoine Rolet.
\newblock Smooth and {Sparse} {Optimal} {Transport}.
\newblock In \emph{{AISTATS}}, 2018.

\bibitem[Bousquet et~al.(2017)Bousquet, Gelly, Tolstikhin, Simon-Gabriel, and
  Schoelkopf]{bousquet_optimal_2017}
Olivier Bousquet, Sylvain Gelly, Ilya Tolstikhin, Carl-Johann Simon-Gabriel,
  and Bernhard Schoelkopf.
\newblock From optimal transport to generative modeling: the {VEGAN} cookbook.
\newblock \emph{arXiv}, 1705.07642, 2017.

\bibitem[Bunke \& Shearer(1998)Bunke and Shearer]{bunke_graph_1998}
Horst Bunke and Kim Shearer.
\newblock A graph distance metric based on the maximal common subgraph.
\newblock \emph{Pattern Recognition Letters}, 19\penalty0 (3):\penalty0
  255--259, 1998.

\bibitem[Canino et~al.(1998)Canino, Ottusch, Stalzer, Visher, and
  Wandzura]{canino_numerical_1998}
Lawrence~F. Canino, John~J. Ottusch, Mark~A. Stalzer, John~L. Visher, and
  Stephen~M. Wandzura.
\newblock Numerical {Solution} of the {Helmholtz} {Equation} in {2D} and {3D}
  {Using} a {High}-{Order} {Nyström} {Discretization}.
\newblock \emph{Journal of Computational Physics}, 146\penalty0 (2):\penalty0
  627--663, 1998.

\bibitem[Charikar(2002)]{charikar_similarity_2002}
Moses Charikar.
\newblock Similarity estimation techniques from rounding algorithms.
\newblock In \emph{{ACM} symposium on {Theory} of computing ({STOC})}, 2002.

\bibitem[Chizat et~al.(2018)Chizat, Peyré, Schmitzer, and
  Vialard]{chizat_scaling_2018}
Lénaïc Chizat, Gabriel Peyré, Bernhard Schmitzer, and François-Xavier
  Vialard.
\newblock Scaling algorithms for unbalanced optimal transport problems.
\newblock \emph{Mathematics of Computation}, 87\penalty0 (314):\penalty0
  2563--2609, 2018.

\bibitem[Cohn(2017)]{cohn_conceptual_2017}
Henry Cohn.
\newblock A {Conceptual} {Breakthrough} in {Sphere} {Packing}.
\newblock \emph{Notices of the American Mathematical Society}, 64\penalty0
  (02):\penalty0 102--115, 2017.

\bibitem[Conneau et~al.(2018)Conneau, Lample, Ranzato, Denoyer, and
  Jégou]{conneau_word_2018}
Alexis Conneau, Guillaume Lample, Marc'Aurelio Ranzato, Ludovic Denoyer, and
  Hervé Jégou.
\newblock Word translation without parallel data.
\newblock In \emph{{ICLR}}, 2018.

\bibitem[Courty et~al.(2017)Courty, Flamary, Habrard, and
  Rakotomamonjy]{courty_joint_2017}
Nicolas Courty, Rémi Flamary, Amaury Habrard, and Alain Rakotomamonjy.
\newblock Joint distribution optimal transportation for domain adaptation.
\newblock In \emph{{NeurIPS}}, 2017.

\bibitem[Cuturi(2013)]{cuturi_sinkhorn_2013}
Marco Cuturi.
\newblock Sinkhorn {Distances}: {Lightspeed} {Computation} of {Optimal}
  {Transport}.
\newblock In \emph{{NeurIPS}}, 2013.

\bibitem[Ding et~al.(2017)Ding, Kondor, and
  Eskreis-Winkler]{ding_multiresolution_2017}
Yi~Ding, Risi Kondor, and Jonathan Eskreis-Winkler.
\newblock Multiresolution {Kernel} {Approximation} for {Gaussian} {Process}
  {Regression}.
\newblock In \emph{{NeurIPS}}, 2017.

\bibitem[Dvurechensky et~al.(2018)Dvurechensky, Gasnikov, and
  Kroshnin]{dvurechensky_computational_2018}
Pavel~E. Dvurechensky, Alexander Gasnikov, and Alexey Kroshnin.
\newblock Computational {Optimal} {Transport}: {Complexity} by {Accelerated}
  {Gradient} {Descent} {Is} {Better} {Than} by {Sinkhorn}'s {Algorithm}.
\newblock In \emph{{ICML}}, 2018.

\bibitem[Essid \& Solomon(2018)Essid and Solomon]{essid_quadratically_2018}
Montacer Essid and Justin Solomon.
\newblock Quadratically {Regularized} {Optimal} {Transport} on {Graphs}.
\newblock \emph{SIAM Journal on Scientific Computing}, 40\penalty0
  (4):\penalty0 A1961--A1986, 2018.

\bibitem[Fey \& Lenssen(2019)Fey and Lenssen]{fey_fast_2019}
Matthias Fey and Jan~E. Lenssen.
\newblock Fast {Graph} {Representation} {Learning} with {PyTorch} {Geometric}.
\newblock In \emph{Workshop on {Representation} {Learning} on {Graphs} and
  {Manifolds}, {ICLR}}, 2019.

\bibitem[Forrow et~al.(2019)Forrow, Hütter, Nitzan, Rigollet, Schiebinger, and
  Weed]{forrow_statistical_2019}
Aden Forrow, Jan-Christian Hütter, Mor Nitzan, Philippe Rigollet, Geoffrey
  Schiebinger, and Jonathan Weed.
\newblock Statistical {Optimal} {Transport} via {Factored} {Couplings}.
\newblock In \emph{{AISTATS}}, 2019.

\bibitem[Frogner et~al.(2015)Frogner, Zhang, Mobahi, Araya-Polo, and
  Poggio]{frogner_learning_2015}
Charlie Frogner, Chiyuan Zhang, Hossein Mobahi, Mauricio Araya-Polo, and
  Tomaso~A. Poggio.
\newblock Learning with a {Wasserstein} {Loss}.
\newblock In \emph{{NeurIPS}}, 2015.

\bibitem[Gasteiger et~al.(2019)Gasteiger, Bojchevski, and
  Günnemann]{gasteiger_predict_2019}
Johannes Gasteiger, Aleksandar Bojchevski, and Stephan Günnemann.
\newblock Predict then {Propagate}: {Graph} {Neural} {Networks} {Meet}
  {Personalized} {PageRank}.
\newblock In \emph{{ICLR}}, 2019.

\bibitem[Gasteiger et~al.(2020)Gasteiger, Groß, and
  Günnemann]{gasteiger_directional_2020}
Johannes Gasteiger, Janek Groß, and Stephan Günnemann.
\newblock Directional {Message} {Passing} for {Molecular} {Graphs}.
\newblock In \emph{{ICLR}}, 2020.

\bibitem[Genevay et~al.(2016)Genevay, Cuturi, Peyré, and
  Bach]{genevay_stochastic_2016}
Aude Genevay, Marco Cuturi, Gabriel Peyré, and Francis~R. Bach.
\newblock Stochastic {Optimization} for {Large}-scale {Optimal} {Transport}.
\newblock In \emph{{NeurIPS}}, 2016.

\bibitem[Genevay et~al.(2018)Genevay, Peyré, and
  Cuturi]{genevay_learning_2018}
Aude Genevay, Gabriel Peyré, and Marco Cuturi.
\newblock Learning {Generative} {Models} with {Sinkhorn} {Divergences}.
\newblock In \emph{{AISTATS}}, 2018.

\bibitem[Gerber \& Maggioni(2017)Gerber and Maggioni]{gerber_multiscale_2017}
Samuel Gerber and Mauro Maggioni.
\newblock Multiscale {Strategies} for {Computing} {Optimal} {Transport}.
\newblock \emph{J. Mach. Learn. Res.}, 18:\penalty0 72:1--72:32, 2017.

\bibitem[Grave et~al.(2019)Grave, Joulin, and Berthet]{grave_unsupervised_2019}
Edouard Grave, Armand Joulin, and Quentin Berthet.
\newblock Unsupervised {Alignment} of {Embeddings} with {Wasserstein}
  {Procrustes}.
\newblock In \emph{{AISTATS}}, 2019.

\bibitem[Houston et~al.(2019)Houston, Nandi, and Bowman]{houston_machine_2019}
Paul~L. Houston, Apurba Nandi, and Joel~M. Bowman.
\newblock A {Machine} {Learning} {Approach} for {Prediction} of {Rate}
  {Constants}.
\newblock \emph{The Journal of Physical Chemistry Letters}, 10\penalty0
  (17):\penalty0 5250--5258, 2019.

\bibitem[Indyk \& Thaper(2003)Indyk and Thaper]{indyk_fast_2003}
Piotr Indyk and Nitin Thaper.
\newblock Fast image retrieval via embeddings.
\newblock In \emph{International {Workshop} on {Statistical} and
  {Computational} {Theories} of {Vision}, {ICCV}}, 2003.

\bibitem[Jambulapati et~al.(2019)Jambulapati, Sidford, and
  Tian]{jambulapati_direct_2019}
Arun Jambulapati, Aaron Sidford, and Kevin Tian.
\newblock A {Direct} tilde\{{O}\}(1/epsilon) {Iteration} {Parallel} {Algorithm}
  for {Optimal} {Transport}.
\newblock In \emph{{NeurIPS}}, 2019.

\bibitem[Johnson et~al.(2015)Johnson, Krishna, Stark, Li, Shamma, Bernstein,
  and Li]{johnson_image_2015}
Justin Johnson, Ranjay Krishna, Michael Stark, Li-Jia Li, David~A. Shamma,
  Michael~S. Bernstein, and Fei-Fei Li.
\newblock Image retrieval using scene graphs.
\newblock In \emph{{CVPR}}, 2015.

\bibitem[Joulin et~al.(2018)Joulin, Bojanowski, Mikolov, Jégou, and
  Grave]{joulin_loss_2018}
Armand Joulin, Piotr Bojanowski, Tomas Mikolov, Hervé Jégou, and Edouard
  Grave.
\newblock Loss in {Translation}: {Learning} {Bilingual} {Word} {Mapping} with a
  {Retrieval} {Criterion}.
\newblock In \emph{{EMNLP}}, 2018.

\bibitem[Kipf \& Welling(2017)Kipf and Welling]{kipf_semi-supervised_2017}
Thomas~N. Kipf and Max Welling.
\newblock Semi-{Supervised} {Classification} with {Graph} {Convolutional}
  {Networks}.
\newblock In \emph{{ICLR}}, 2017.

\bibitem[Ktena et~al.(2017)Ktena, Parisot, Ferrante, Rajchl, Lee, Glocker, and
  Rueckert]{ktena_distance_2017}
Sofia~Ira Ktena, Sarah Parisot, Enzo Ferrante, Martin Rajchl, Matthew Lee, Ben
  Glocker, and Daniel Rueckert.
\newblock Distance {Metric} {Learning} {Using} {Graph} {Convolutional}
  {Networks}: {Application} to {Functional} {Brain} {Networks}.
\newblock In \emph{{MICCAI}}, 2017.

\bibitem[Lerouge et~al.(2017)Lerouge, Abu-Aisheh, Raveaux, Héroux, and
  Adam]{lerouge_new_2017}
Julien Lerouge, Zeina Abu-Aisheh, Romain Raveaux, Pierre Héroux, and
  Sébastien Adam.
\newblock New binary linear programming formulation to compute the graph edit
  distance.
\newblock \emph{Pattern Recognit.}, 72:\penalty0 254--265, 2017.

\bibitem[Li et~al.(2019)Li, Gu, Dullien, Vinyals, and Kohli]{li_graph_2019}
Yujia Li, Chenjie Gu, Thomas Dullien, Oriol Vinyals, and Pushmeet Kohli.
\newblock Graph {Matching} {Networks} for {Learning} the {Similarity} of
  {Graph} {Structured} {Objects}.
\newblock In \emph{{ICML}}, 2019.

\bibitem[Maretic et~al.(2019)Maretic, Gheche, Chierchia, and
  Frossard]{maretic_got_2019}
Hermina~Petric Maretic, Mireille~El Gheche, Giovanni Chierchia, and Pascal
  Frossard.
\newblock {GOT}: {An} {Optimal} {Transport} framework for {Graph} comparison.
\newblock In \emph{{NeurIPS}}, 2019.

\bibitem[Meng et~al.(2019)Meng, Ke, Zhang, Zhang, Zhong, and
  Ma]{meng_large-scale_2019}
Cheng Meng, Yuan Ke, Jingyi Zhang, Mengrui Zhang, Wenxuan Zhong, and Ping Ma.
\newblock Large-scale optimal transport map estimation using projection
  pursuit.
\newblock In \emph{{NeurIPS}}, 2019.

\bibitem[Mensch \& Peyré(2020)Mensch and Peyré]{mensch_online_2020}
Arthur Mensch and Gabriel Peyré.
\newblock Online {Sinkhorn}: {Optimal} {Transport} distances from sample
  streams.
\newblock In \emph{{NeurIPS}}, 2020.

\bibitem[Musco \& Musco(2017)Musco and Musco]{musco_recursive_2017}
Cameron Musco and Christopher Musco.
\newblock Recursive {Sampling} for the {Nystrom} {Method}.
\newblock In \emph{{NeurIPS}}, 2017.

\bibitem[Neuhaus \& Bunke(2004)Neuhaus and Bunke]{neuhaus_error-tolerant_2004}
Michel Neuhaus and Horst Bunke.
\newblock An {Error}-{Tolerant} {Approximate} {Matching} {Algorithm} for
  {Attributed} {Planar} {Graphs} and {Its} {Application} to {Fingerprint}
  {Classification}.
\newblock In \emph{Structural, {Syntactic}, and {Statistical} {Pattern}
  {Recognition}}, 2004.

\bibitem[Nikolentzos et~al.(2017)Nikolentzos, Meladianos, and
  Vazirgiannis]{nikolentzos_matching_2017}
Giannis Nikolentzos, Polykarpos Meladianos, and Michalis Vazirgiannis.
\newblock Matching {Node} {Embeddings} for {Graph} {Similarity}.
\newblock In \emph{{AAAI}}, 2017.

\bibitem[Nistér \& Stewénius(2006)Nistér and
  Stewénius]{nister_scalable_2006}
David Nistér and Henrik Stewénius.
\newblock Scalable {Recognition} with a {Vocabulary} {Tree}.
\newblock In \emph{{CVPR}}, 2006.

\bibitem[Papadakis et~al.(2014)Papadakis, Peyré, and
  Oudet]{papadakis_optimal_2014}
Nicolas Papadakis, Gabriel Peyré, and Édouard Oudet.
\newblock Optimal {Transport} with {Proximal} {Splitting}.
\newblock \emph{SIAM J. Imaging Sciences}, 7\penalty0 (1):\penalty0 212--238,
  2014.

\bibitem[Paszke et~al.(2019)Paszke, Gross, Massa, Lerer, Bradbury, Chanan,
  Killeen, Lin, Gimelshein, Antiga, Desmaison, Köpf, Yang, DeVito, Raison,
  Tejani, Chilamkurthy, Steiner, Fang, Bai, and Chintala]{paszke_pytorch_2019}
Adam Paszke, Sam Gross, Francisco Massa, Adam Lerer, James Bradbury, Gregory
  Chanan, Trevor Killeen, Zeming Lin, Natalia Gimelshein, Luca Antiga, Alban
  Desmaison, Andreas Köpf, Edward Yang, Zachary DeVito, Martin Raison, Alykhan
  Tejani, Sasank Chilamkurthy, Benoit Steiner, Lu~Fang, Junjie Bai, and Soumith
  Chintala.
\newblock {PyTorch}: {An} {Imperative} {Style}, {High}-{Performance} {Deep}
  {Learning} {Library}.
\newblock In \emph{{NeurIPS}}, 2019.

\bibitem[Paulevé et~al.(2010)Paulevé, Jégou, and
  Amsaleg]{pauleve_locality_2010}
Loïc Paulevé, Hervé Jégou, and Laurent Amsaleg.
\newblock Locality sensitive hashing: {A} comparison of hash function types and
  querying mechanisms.
\newblock \emph{Pattern Recognit. Lett.}, 31\penalty0 (11):\penalty0
  1348--1358, 2010.

\bibitem[Percus \& Martin(1998)Percus and Martin]{percus_scaling_1998}
Allon~G Percus and Olivier~C Martin.
\newblock Scaling {Universalities} ofkth-{Nearest} {Neighbor} {Distances} on
  {Closed} {Manifolds}.
\newblock \emph{Advances in Applied Mathematics}, 21\penalty0 (3):\penalty0
  424--436, 1998.

\bibitem[Peyré \& Cuturi(2019)Peyré and Cuturi]{peyre_computational_2019}
Gabriel Peyré and Marco Cuturi.
\newblock Computational {Optimal} {Transport}.
\newblock \emph{Foundations and Trends in Machine Learning}, 11\penalty0
  (5-6):\penalty0 355--607, 2019.

\bibitem[Rabin et~al.(2011)Rabin, Peyré, Delon, and
  Bernot]{rabin_wasserstein_2011}
Julien Rabin, Gabriel Peyré, Julie Delon, and Marc Bernot.
\newblock Wasserstein {Barycenter} and {Its} {Application} to {Texture}
  {Mixing}.
\newblock In \emph{Scale {Space} and {Variational} {Methods} in {Computer}
  {Vision} ({SSVM})}, 2011.

\bibitem[Riba et~al.(2018)Riba, Fischer, Lladós, and
  Fornés]{riba_learning_2018}
Pau Riba, Andreas Fischer, Josep Lladós, and Alicia Fornés.
\newblock Learning {Graph} {Distances} with {Message} {Passing} {Neural}
  {Networks}.
\newblock In \emph{{ICPR}}, 2018.

\bibitem[Riesen \& Bunke(2008)Riesen and Bunke]{riesen_iam_2008}
Kaspar Riesen and Horst Bunke.
\newblock {IAM} {Graph} {Database} {Repository} for {Graph} {Based} {Pattern}
  {Recognition} and {Machine} {Learning}.
\newblock In \emph{Structural, {Syntactic}, and {Statistical} {Pattern}
  {Recognition}}, 2008.

\bibitem[Riesen \& Bunke(2009)Riesen and Bunke]{riesen_approximate_2009}
Kaspar Riesen and Horst Bunke.
\newblock Approximate graph edit distance computation by means of bipartite
  graph matching.
\newblock \emph{Image Vis. Comput.}, 27\penalty0 (7):\penalty0 950--959, 2009.

\bibitem[Ruder et~al.(2019)Ruder, Vulic, and Søgaard]{ruder_survey_2019}
Sebastian Ruder, Ivan Vulic, and Anders Søgaard.
\newblock A {Survey} of {Cross}-lingual {Word} {Embedding} {Models}.
\newblock \emph{J. Artif. Intell. Res.}, 65:\penalty0 569--631, 2019.

\bibitem[Schmitzer(2019)]{schmitzer_stabilized_2019}
Bernhard Schmitzer.
\newblock Stabilized {Sparse} {Scaling} {Algorithms} for {Entropy}
  {Regularized} {Transport} {Problems}.
\newblock \emph{SIAM Journal on Scientific Computing}, 41\penalty0
  (3):\penalty0 A1443--A1481, 2019.

\bibitem[Shrivastava \& Li(2014)Shrivastava and
  Li]{shrivastava_asymmetric_2014}
Anshumali Shrivastava and Ping Li.
\newblock Asymmetric {LSH} ({ALSH}) for {Sublinear} {Time} {Maximum} {Inner}
  {Product} {Search} ({MIPS}).
\newblock In \emph{{NeurIPS}}, 2014.

\bibitem[Si et~al.(2017)Si, Hsieh, and Dhillon]{si_memory_2017}
Si~Si, Cho-Jui Hsieh, and Inderjit~S. Dhillon.
\newblock Memory {Efficient} {Kernel} {Approximation}.
\newblock \emph{Journal of Machine Learning Research}, 18\penalty0
  (20):\penalty0 1--32, 2017.

\bibitem[Sinkhorn \& Knopp(1967)Sinkhorn and Knopp]{sinkhorn_concerning_1967}
Richard Sinkhorn and Paul Knopp.
\newblock Concerning nonnegative matrices and doubly stochastic matrices.
\newblock \emph{Pacific Journal of Mathematics}, 21\penalty0 (2):\penalty0
  343--348, 1967.

\bibitem[Solomon et~al.(2014)Solomon, Rustamov, Guibas, and
  Butscher]{solomon_earth_2014}
Justin Solomon, Raif~M. Rustamov, Leonidas~J. Guibas, and Adrian Butscher.
\newblock Earth mover's distances on discrete surfaces.
\newblock \emph{ACM Trans. Graph.}, 33\penalty0 (4):\penalty0 67:1--67:12,
  2014.

\bibitem[Solomon et~al.(2015)Solomon, Goes, Peyré, Cuturi, Butscher, Nguyen,
  Du, and Guibas]{solomon_convolutional_2015}
Justin Solomon, Fernando~de Goes, Gabriel Peyré, Marco Cuturi, Adrian
  Butscher, Andy Nguyen, Tao Du, and Leonidas~J. Guibas.
\newblock Convolutional wasserstein distances: efficient optimal transportation
  on geometric domains.
\newblock \emph{ACM Trans. Graph.}, 34\penalty0 (4):\penalty0 66:1--66:11,
  2015.

\bibitem[Stanford(2014)]{stanford_stanford_2014}
Computer Graphics~Laboratory Stanford.
\newblock The {Stanford} {3D} {Scanning} {Repository}, 2014.
\newblock URL \url{http://graphics.stanford.edu/data/3Dscanrep/}.

\bibitem[Tarjan(1997)]{tarjan_dynamic_1997}
Robert~E. Tarjan.
\newblock Dynamic trees as search trees via euler tours, applied to the network
  simplex algorithm.
\newblock \emph{Mathematical Programming}, 78\penalty0 (2):\penalty0 169--177,
  1997.

\bibitem[Tenetov et~al.(2018)Tenetov, Wolansky, and Kimmel]{tenetov_fast_2018}
Evgeny Tenetov, Gershon Wolansky, and Ron Kimmel.
\newblock Fast {Entropic} {Regularized} {Optimal} {Transport} {Using}
  {Semidiscrete} {Cost} {Approximation}.
\newblock \emph{SIAM J. Sci. Comput.}, 40\penalty0 (5):\penalty0 A3400--A3422,
  2018.

\bibitem[Vaswani et~al.(2017)Vaswani, Shazeer, Parmar, Uszkoreit, Jones, Gomez,
  Kaiser, and Polosukhin]{vaswani_attention_2017}
Ashish Vaswani, Noam Shazeer, Niki Parmar, Jakob Uszkoreit, Llion Jones,
  Aidan~N. Gomez, Lukasz Kaiser, and Illia Polosukhin.
\newblock Attention is {All} you {Need}.
\newblock In \emph{{NeurIPS}}, 2017.

\bibitem[Vayer et~al.(2019)Vayer, Courty, Tavenard, Chapel, and
  Flamary]{vayer_optimal_2019}
Titouan Vayer, Nicolas Courty, Romain Tavenard, Laetitia Chapel, and Rémi
  Flamary.
\newblock Optimal {Transport} for structured data with application on graphs.
\newblock In \emph{{ICML}}, 2019.

\bibitem[Wang et~al.(2014)Wang, Shen, Song, and Ji]{wang_hashing_2014}
Jingdong Wang, Heng~Tao Shen, Jingkuan Song, and Jianqiu Ji.
\newblock Hashing for {Similarity} {Search}: {A} {Survey}.
\newblock \emph{arXiv}, 1408.2927, 2014.

\bibitem[Wang et~al.(2019)Wang, Yan, and Yang]{wang_learning_2019}
Runzhong Wang, Junchi Yan, and Xiaokang Yang.
\newblock Learning {Combinatorial} {Embedding} {Networks} for {Deep} {Graph}
  {Matching}.
\newblock In \emph{{ICCV}}, 2019.

\bibitem[Williams \& Seeger(2001)Williams and Seeger]{williams_using_2001}
Christopher K.~I. Williams and Matthias Seeger.
\newblock Using the {Nyström} {Method} to {Speed} {Up} {Kernel} {Machines}.
\newblock In \emph{{NeurIPS}}, 2001.

\bibitem[Xiao et~al.(2008)Xiao, Gao, Tao, and Li]{xiao_hmm-based_2008}
Bing Xiao, Xinbo Gao, Dacheng Tao, and Xuelong Li.
\newblock {HMM}-based graph edit distance for image indexing.
\newblock \emph{Int. J. Imaging Systems and Technology}, 18\penalty0
  (2-3):\penalty0 209--218, 2008.

\bibitem[Zambaldi et~al.(2019)Zambaldi, Raposo, Santoro, Bapst, Li, Babuschkin,
  Tuyls, Reichert, Lillicrap, Lockhart, Shanahan, Langston, Pascanu, Botvinick,
  Vinyals, and Battaglia]{zambaldi_deep_2019}
Vinícius~Flores Zambaldi, David Raposo, Adam Santoro, Victor Bapst, Yujia Li,
  Igor Babuschkin, Karl Tuyls, David~P. Reichert, Timothy~P. Lillicrap, Edward
  Lockhart, Murray Shanahan, Victoria Langston, Razvan Pascanu, Matthew
  Botvinick, Oriol Vinyals, and Peter~W. Battaglia.
\newblock Deep reinforcement learning with relational inductive biases.
\newblock In \emph{{ICLR}}, 2019.

\bibitem[Zhang et~al.(2008)Zhang, Tsang, and Kwok]{zhang_improved_2008}
Kai Zhang, Ivor~W. Tsang, and James~T. Kwok.
\newblock Improved {Nyström} low-rank approximation and error analysis.
\newblock In \emph{{ICML}}, 2008.

\end{thebibliography}
\bibliographystyle{fast-ot}


\appendix
\setcounter{theorem}{0}
\renewcommand*{\thetheorem}{\Alph{theorem}}
\setcounter{lemma}{0}
\renewcommand*{\thelemma}{\Alph{lemma}}

\section{Complexity analysis} \label{app:complexity}

\textbf{Sparse Sinkhorn.} A common way of achieving a high $p_1$ and low $p_2$ in LSH is via the AND-OR construction. In this scheme we calculate $B \cdot r$ hash functions, divided into $B$ sets (hash bands) of $r$ hash functions each. A pair of points is considered as neighbors if any hash band matches completely. Calculating the hash buckets for all points with $b$ hash buckets per function scales as $\mathcal{O}((n + m) d B b r)$ for the hash functions we consider. As expected, for the tasks and hash functions we investigated we obtain approximately $m / b^r$ and $n / b^r$ neighbors, with $b^r$ hash buckets per band. Using this we can fix the number of neighbors to a small, constant $\beta$ in expectation with $b^r = \min(n, m) / \beta$. We thus obtain a sparse cost matrix $\mC^\text{sp}$ with $\mathcal{O}(\max(n, m) \beta)$ non-infinite values and can calculate $\vs$ and $\vt$ in linear time $\mathcal{O}(N_\text{sink} \max(n, m) \beta)$, where $N_\text{sink} \le 2 + \frac{-4 \ln (\min_{i,j} \{ \tilde{\mK}_{ij} | \tilde{\mK}_{ij} > 0 \} \min_{i,j} \{ \vp_i, \vq_j \} )}{\varepsilon}$ (see \cref{th:convergence}) denotes the number of Sinkhorn iterations. Calculating the hash buckets with $r = \frac{\log \min(n, m) - \log \beta}{\log b}$ takes $\mathcal{O}((n + m) d B b (\log \min(n, m) - \log \beta) / \log b)$. Since $B$, $b$, and $\beta$ are small, we obtain roughly log-linear scaling with the number of points overall, i.e.\ $\mathcal{O}(n \log n)$ for $n \approx m$.

\textbf{LCN-Sinkhorn.} Both choosing landmarks via $k$-means++ sampling and via $k$-means with a fixed number of iterations have the same runtime complexity of $\mathcal{O}((n + m)ld)$. Precomputing $\mW$ can be done in time $\mathcal{O}(nl^2 + l^3)$. The low-rank part of updating the vectors $\vs$ and $\vt$ can be computed in $\mathcal{O}(nl + l^2 + lm)$, with $l$ chosen constant, i.e.\ independently of $n$ and $m$. Since sparse Sinkhorn with LSH has a log-linear runtime we again obtain log-linear overall runtime for LCN-Sinkhorn.

\section{Limitations} \label{app:limitations}

\textbf{Sparse Sinkhorn.} Using a sparse approximation for $\mK$ works well in the common case when the regularization parameter $\lambda$ is low and the cost function varies enough between data pairs, such that the transport plan $\mP$ resembles a sparse matrix. However, it can fail if the cost between pairs is very similar or the regularization is very high, if the dataset contains many hubs, i.e.\ points with a large number of neighbors, or if the distributions $\vp$ or $\vq$ are spread very unevenly. Furthermore, sparse Sinkhorn can be too unstable to train a model from scratch, since randomly initialized embeddings often have no close neighbors (see \cref{sec:exp}). Note also that LSH requires the cost function to be associated with a metric space, while regular Sinkhorn can be used with arbitrary costs.

Note that we are only interested in an approximate solution with finite error $\varepsilon$. We therefore do not need the kernel matrix to be fully indecomposable or have total support, which would be necessary and sufficient for a unique (up to a scalar factor) and exact solution, respectively \citep{sinkhorn_concerning_1967}. However, the sparse approximation is not guaranteed to have support (\cref{def:support}), which is necessary and sufficient for the Sinkhorn algorithm to converge. The approximated matrix is actually very likely not to have support if we use one LSH bucket per sample. This is due to the non-quadratic block structure resulting from every point only having non-zero entries for points in the other data set that fall in the same bucket. We can alleviate this problem by using unbalanced OT, as proposed in \cref{sec:gtn}, or (empirically) the AND-OR construction. We can also simply choose to ignore this as long as we limit the maximum number of Sinkhorn iterations. On the 3D point cloud and random data experiments we indeed ignored this issue and actually observed good performance. Experiments with other LSH schemes and the AND-OR construction showed no performance improvement despite the associated cost matrices having support. Not having support therefore seems not to be an issue in practice, at least for the data we investigated.

\textbf{LCN-Sinkhorn.} The LCN approximation is guaranteed to have support due to the Nyström part. Other weak spots of sparse Sinkhorn, such as very similar cost between pairs, high regularization, or data containing many hubs, are also usually handled well by the Nyström part of LCN.
Highly concentrated distributions $\vp$ and $\vq$ can still have adverse effects on LCN-Sinkhorn. We can compensate for these by sampling landmarks or neighbors proportional to each point's probability mass.

The Nyström part of LCN also has its limits, though. If the regularization parameter is low or the cost function varies greatly, we observed stability issues (over- and underflows) of the Nyström approximation because of the inverse $\mA^{-1}$, which cannot be calculated in log-space. The Nyström approximation furthermore is not guaranteed to be non-negative, which can lead to catastrophic failures if the matrix product in \cref{eq:sinkhorn} becomes negative. In these extreme cases we also observed catastrophic elimination with the correction $\mK^\text{sp}_\Delta$.
Since a low entropy regularization essentially means that optimal transport is very local, we recommend using sparse Sinkhorn in these scenarios. This again demonstrates the complementarity of the sparse approximation and Nyström: In cases where one fails we can often resort to the other.

\section{Proof of \cref{th:uniform_error}} \label{app:uniform_error}

By linearity of expectation we obtain
\begin{equation}
\begin{aligned}
    \E[ \mK_{i, i_k} - \mK_{\text{Nys}, i, i_k} ] &= \E[\mK_{i, i_k}] - \E[\mK_{\text{Nys}, i, i_k}]\\
    &= \E[e^{-\delta_k/\lambda}] - \E[\mK_{\text{Nys}, i, i_k}]
\end{aligned}
\end{equation}
with the distance to the $k$th-nearest neighbor $\delta_k$. Note that without loss of generality we can assume unit manifold volume and obtain the integral resulting from the first expectation as (ignoring boundary effects that are exponentially small in $n$, see \citet{percus_scaling_1998})
\begin{equation}
    \E[e^{-\delta_k/\lambda}] \approx \frac{n!}{(n-k)!(k-1)!} \int_0^{\frac{((d/2)!)^{1/d}}{\sqrt{\pi}}} e^{-r/\lambda} V_d(r)^{k-1} (1-V_d(r))^{n-k} \frac{\partial V_d(r)}{\partial r} \,\textnormal{d}r,
\end{equation}
with the volume of the $d$-ball
\begin{equation}
    V_d(r) = \frac{\pi^{d/2}r^d}{(d/2)!}.
\end{equation}
Since this integral does not have an analytical solution we can either calculate it numerically or lower bound it using Jensen's inequality (again ignoring exponentially small boundary effects)
\begin{equation}
    \E[e^{-\delta_k/\lambda}] \ge e^{-\E[\delta_k]/\lambda} \approx \exp \left( -\frac{((d/2)!)^{1/d}}{\sqrt{\pi} \lambda} \frac{(k-1 + 1/d)!}{(k-1)!} \frac{n!}{(n + 1/d)!} \right).
\end{equation}

To upper bound the second expectation $\E[\mK_{\text{Nys}, i, i_k}]$ we now denote the distance between two points by $r_{ia} = \|\vx_{\text{p}i} - \vx_a\|_2$, the kernel by $k_{ia} = e^{-r_{ia} / \lambda}$ and the inter-landmark kernel matrix by $\mK_{\text{L}}$.  We first consider
\begin{equation}
\begin{aligned}
    p(\vx_j &\mid \vx_j \text{ is $k$th-nearest neighbor}) =\\
    &= \int p(\delta_k = r_{ij} \mid \vx_i, \vx_j) p(\vx_i) p(\vx_j) \,\text{d}\vx_i\\
    &= \int p(\delta_k = r_{ij} \mid r_{ij}) p(r_{ij} \mid \vx_j) \,\text{d}r_{ij} \, p(\vx_j)\\
    &= \int p(\delta_k = r_{ij} \mid r_{ij}) p(r_{ij}) \,\text{d}r_{ij} \, p(\vx_j)\\
    &= \int p(\delta_k = r_{ij}) \,\text{d}r_{ij} \, p(\vx_j)\\
    &= p(\vx_j) = p(\vx_i),
\end{aligned}
\label{eq:knearest}
\end{equation}
where the third step is due to the uniform distribution. Since landmarks are more than $2R$ apart we can approximate
\begin{equation}
\begin{aligned}
    \mK_{\text{L}}^{-1} = (I_l + \vone_{l \times l} \gO(e^{-2R/\lambda}))^{-1} = I_n - \vone_{l \times l} \gO(e^{-2R/\lambda}),
\end{aligned}
\label{eq:K_inverse}
\end{equation}
where $\vone_{l \times l}$ denotes the constant 1 matrix, with the number of landmarks $l$.
We can now use (1) the fact that landmarks are arranged apriori, (2) Hölder's inequality, (3) \cref{eq:knearest}, and (4) \cref{eq:K_inverse} to obtain
\begin{equation}
\begin{aligned}
    \E[\mK_{\text{Nys}, i, i_k}] &= \E\left[\sum_{a=1}^l \sum_{b=1}^l k_{ia} (\mK_{\text{L}}^{-1})_{ab} k_{i_kb}\right]\\
    &\overset{(1)}{=} \sum_{a=1}^l \sum_{b=1}^l (\mK_{\text{L}}^{-1})_{ab} \E[k_{ia} k_{i_kb}]\\
    &\overset{(2)}{\le} \sum_{a=1}^l \sum_{b=1}^l (\mK_{\text{L}}^{-1})_{ab} \E[k_{ia}^2]^{1/2} \E[k_{i_kb}^2]^{1/2}\\
    &\overset{(3)}{=} \sum_{a=1}^l \sum_{b=1}^l (\mK_{\text{L}}^{-1})_{ab} \E[k_{ia}^2]^{1/2} \E[k_{ib}^2]^{1/2}\\
    &\overset{(4)}{=} \sum_{a=1}^l \E[k_{ia}^2] - \gO(e^{-2R/\lambda}).
\end{aligned}
\end{equation}
Since landmarks are more than $2R$ apart we have $V_{\gM} \ge l V_d(R)$, where $V_{\gM}$ denotes the volume of the manifold. Assuming Euclideanness in $V_d(R)$ we can thus use the fact that data points are uniformly distributed to obtain
\begin{equation}
\begin{split}
    \E[k_{ia}^2] &= \E[e^{-2r_{ia}/\lambda}]\\
    &= \frac{1}{V_{\gM}} \int e^{-2r/\lambda} \frac{\partial V_d(r)}{\partial r} \,\textnormal{d}r\\
    &\le \frac{1}{l V_d(R)} \int e^{-2r/\lambda} \frac{\partial V_d(r)}{\partial r} \,\textnormal{d}r\\
    &= \frac{1}{l V_d(R)} \int_0^R e^{-2r/\lambda} \frac{\partial V_d(r)}{\partial r} \,\textnormal{d}r + \gO(e^{-2R/\lambda})\\
    &= \frac{d}{lR^d} \int_0^R e^{-2r/\lambda} r^{d-1} \,\textnormal{d}r + \gO(e^{-2R/\lambda})\\
    &= \frac{d\left( \Gamma(d) - \Gamma(d, 2R/\lambda) \right)}{l(2R/\lambda)^d} + \gO(e^{-2R/\lambda})
\end{split}
\end{equation}
and finally
\begin{equation}
\begin{aligned}
    \E[\mK_{\text{Nys}, i, i_k}]
    \le {}& \sum_{a=1}^l \E[k_{ia}^2] - \gO(e^{-2R/\lambda})\\
    \le {}& \frac{d\left( \Gamma(d) - \Gamma(d, 2R/\lambda) \right)}{(2R/\lambda)^d} + \gO(e^{-2R/\lambda}).
\end{aligned}
\end{equation}\qed

\section{Proof of \cref{th:cluster_error}} \label{app:cluster_error}

We first prove two lemmas that will be useful later on.

\begin{lemma}
\label{lem:nystrombound}
Let $\tilde{\mK}$ be the Nyström approximation of the similarity matrix $\mK_{ij} = e^{-\|\vx_i - \vx_j\|_2/\lambda}$, with all Nyström landmarks being at least $D$ apart and data samples being no more than $r$ away from its closest landmark. Then
\begin{equation}
    \tilde{\mK}_{ij} = \tilde{\mK}_{ij}^{\textnormal{2L}} + \gO(e^{-2\max(D-r, D/2)/\lambda}),
\end{equation}
where $\tilde{\mK}^{\textnormal{2L}}$ denotes the Nyström approximation using only the two landmarks closest to the points $\vx_i$ and $\vx_j$.
\end{lemma}
\textit{Proof.} We denote the landmarks closest to the two points $i$ and $j$ with the indices $a$ and $b$, or jointly with $\sA$, and all other landmarks with $\sC$. We furthermore denote the kernel between the point $i$ and the point $a$ as $k_{ia} = e^{-\|\vx_a - \vx_j\|_2/\lambda}$ and the vector of kernels between a set of points $\sA$ and a point $i$ as $\vk_{\sA i}$.

We can split up $\mA^{-1}$ used in the Nyström approximation
\begin{equation}
    \tilde{\mK} = \mU \mA^{-1} \mV,
\end{equation}
where $\mA_{cd} = k_{cd}$, $\mU_{ic} = k_{ic}$, and $\mV_{dj} = k_{dj}$, into relevant blocks via
\begin{equation}
\begin{split}
    \mA^{-1} &= \begin{pmatrix}
    \mA_{\text{2L}} & \mB\\
    \mB^T & \mA_{\text{other}}
    \end{pmatrix}^{-1}\\
    &= \begin{pmatrix}
    \mA_{\text{2L}}^{-1} + \mA_{\text{2L}}^{-1} \mB (\mA/\mA_{\text{2L}})^{-1} \mB^T \mA_{\text{2L}}^{-1} & -\mA_{\text{2L}}^{-1} \mB (\mA/\mA_{\text{2L}})^{-1}\\
    -(\mA/\mA_{\text{2L}})^{-1} \mB^T \mA_{\text{2L}}^{-1} & (\mA/\mA_{\text{2L}})^{-1}
    \end{pmatrix},
\end{split}
\end{equation}
where $\mA/\mA_{\text{2L}} = \mA_{\text{other}} - \mB^T \mA_{\text{2L}}^{-1} \mB$ denotes the Schur complement. We can thus write the entries of the Nyström approximation as
\begin{equation}
\begin{aligned}
    \tilde{\mK}_{ij} = {} & \vk_{\sA i}^T \mA_{\text{2L}}^{-1} \vk_{\sA j}\\
    & + \vk_{\sA i}^T \mA_{\text{2L}}^{-1} \mB (\mA/\mA_{\text{2L}})^{-1} \mB^T \mA_{\text{2L}}^{-1} \vk_{\sA j}\\
    & - \vk_{\sA i}^T \mA_{\text{2L}}^{-1} \mB (\mA/\mA_{\text{2L}})^{-1} \vk_{\sC j}\\
    & - \vk_{\sC i}^T (\mA/\mA_{\text{2L}})^{-1} \mB^T \mA_{\text{2L}}^{-1} \vk_{\sA j}\\
    & + \vk_{\sC i}^T (\mA/\mA_{\text{2L}})^{-1} \vk_{\sC j}\\
    = {} & \tilde{\mK}_{ij}^{\textnormal{2L}} + (\vk_{\sC i}^T - \vk_{\sA i}^T \mA_{\text{2L}}^{-1} \mB)\\
    &(\mA_{\text{other}} - \mB^T \mA_{\text{2L}}^{-1} \mB)^{-1}\\
    &(\vk_{\sC j} - \mB^T \mA_{\text{2L}}^{-1} \vk_{\sA j}).
\end{aligned}
\label{eq:nys2_error}
\end{equation}
Interestingly, the difference to $\tilde{\mK}_{ij}^{\textnormal{2L}}$ is again a Nyström approximation where each factor is the difference between the correct kernel (e.g.\ $\vk_{\sC j}$) and the previous Nyström approximation of this kernel (e.g.\ $\mB^T \mA_{\text{2L}}^{-1} \vk_{\sA j}$).

We next bound the inverse, starting with
\begin{equation}
\begin{aligned}
\mB^T \mA_{\text{2L}}^{-1} \mB &= \begin{pmatrix}
\vk_{\sC a} & \vk_{\sC b}
\end{pmatrix} \frac{1}{1 - k_{ab}^2} \begin{pmatrix}
1 & -k_{ab}\\
-k_{ab} & 1
\end{pmatrix} \begin{pmatrix}
\vk_{\sC a}^T \\ \vk_{\sC b}^T
\end{pmatrix}\\
&= \frac{1}{1 - k_{ab}^2} \left( \vk_{\sC a} \vk_{\sC a}^T - k_{ab} \vk_{\sC a} \vk_{\sC b}^T - k_{ab} \vk_{\sC b} \vk_{\sC a}^T + \vk_{\sC b} \vk_{\sC b}^T \right)\\
&= \vone_{l-2 \times l-2} (1 + \gO(e^{-2D/\lambda})) \cdot 4 \gO(e^{-2D/\lambda})\\
&= \vone_{l-2 \times l-2} \gO(e^{-2D/\lambda}),
\end{aligned}
\end{equation}
where $\vone_{l-2 \times l-2}$ denotes the constant 1 matrix, with the number of landmarks $l$. The last steps use the fact that landmarks are more than $D$ apart and $0 \le k \le 1$ for all k. For this reason we also have $\mA_{\text{other}} = I_{l-2} + \vone_{l-2 \times l-2} \gO(e^{-D/\lambda})$ and can thus use the Neumann series to obtain
\begin{equation}
\begin{aligned}
(\mA_{\text{other}} - \mB^T \mA_{\text{2L}}^{-1} \mB)^{-1} &= (I_{l-2} + \vone_{l-2 \times l-2} \gO(e^{-D/\lambda}))^{-1}\\
&= I_{l-2} - \vone_{l-2 \times l-2} \gO(e^{-D/\lambda}).
\end{aligned}
\end{equation}
We can analogously bound the other terms in \cref{eq:nys2_error} to obtain
\begin{equation}
\begin{aligned}
    \tilde{\mK}_{ij} = {} & \tilde{\mK}_{ij}^{\textnormal{2L}} + (\vk_{\sC i}^T - \vone_{1 \times l-2} \gO(e^{-D/\lambda}))\\
    &(I_{l-2} - \vone_{l-2 \times l-2} \gO(e^{-D/\lambda}))\\
    &(\vk_{\sC j} - \vone_{l-2 \times 1} \gO(e^{-D/\lambda}))\\
    \overset{(1)}{=} {} & \tilde{\mK}_{ij}^{\textnormal{2L}} + \vk_{\sC i}^T \vk_{\sC j} + \gO(e^{-(D + \max(D-r, D/2))/\lambda})\\
    = {} & \tilde{\mK}_{ij}^{\textnormal{2L}} + \sum_{\substack{1 \le k \le l \\ k \neq a, b}} e^{-(\|\vx_i - \vx_k\|_2 + \|\vx_k - \vx_j\|_2)/\lambda}\\
    &+ \gO(e^{-(D + \max(D-r, D/2))/\lambda})\\
    \overset{(2)}{\le} {} & \tilde{\mK}_{ij}^{\textnormal{2L}} + d e^{-2\max(D-r, D/2)/\lambda}\\
    &+ \gO(e^{-\max(2(D-r), (1+\sqrt{3}) D/2)/\lambda})\\
    = {}& \tilde{\mK}_{ij}^{\textnormal{2L}} + \gO(e^{-2\max(D-r, D/2)/\lambda}),
\end{aligned}
\end{equation}
where $d$ denotes the dimension of $\vx$. Step (1) follows from the fact that any points' second closest landmarks must be at least $\max(D-r, D/2)$ away (since landmarks are at least $D$ apart). This furthermore means that any point can have at most $d$ second closest landmarks at this distance, which we used in step (2).\qed

\begin{lemma}
\label{lem:nystrompair}
Let $\tilde{\mK}$ be the Nyström approximation of the similarity matrix $\mK_{ij} = e^{-\|\vx_i - \vx_j\|_2/\lambda}$. Let $\vx_i$ and $\vx_j$ be data points with equal $L_2$ distance $r_i$ and $r_j$ to all $l$ landmarks, which have the same distance $\Delta>0$ to each other. Then
\upshape
\begin{equation}
    \tilde{\mK}_{ij} = \frac{le^{-(r_i + r_j)/\lambda}}{1 + (l-1)e^{-\Delta/\lambda}}
\end{equation}
\itshape
\end{lemma}
\textit{Proof.} The inter-landmark distance matrix is
\begin{equation}
    \mA = e^{-\Delta/\lambda} \vone_{l \times l} + (1 - e^{-\Delta/\lambda}) \mI_l,
\end{equation}
where $\vone_{l \times l}$ denotes the constant 1 matrix. Using the identity
\begin{equation}
    (b \vone_{n \times n} + (a - b) \mI_n)^{-1} = \frac{-b}{(a-b)(a + (n-1) b)} \vone_{n \times n} + \frac{1}{a - b} \mI_n
\end{equation}
we can compute
\begin{equation}
\begin{split}
    \tilde{\mK}_{ij} &= \mU_{i,:} \mA^{-1} \mV_{:,j}\\
    &=  \begin{pmatrix}
e^{-r_i/\lambda} & e^{-r_i/\lambda} & \cdots
\end{pmatrix} \left( \frac{-e^{-\Delta/\lambda}}{(1-e^{-\Delta/\lambda})(1 + (l - 1) e^{-\Delta/\lambda})} \vone_{l \times l} + \frac{1}{1 - e^{-\Delta/\lambda}} \mI_l \right) \begin{pmatrix}
e^{-r_j/\lambda}\\
e^{-r_j/\lambda}\\
\vdots
\end{pmatrix}\\
&= \frac{e^{-(r_i + r_j)/\lambda}}{1 - e^{-\Delta/\lambda}}  \left( \frac{-l^2 e^{-\Delta/\lambda}}{1 + (l - 1) e^{-\Delta/\lambda}} + l \right)
= \frac{e^{-(r_i + r_j)/\lambda}}{1 - e^{-\Delta/\lambda}} \frac{l - le^{-\Delta/\lambda}}{1 + (l - 1) e^{-\Delta/\lambda}}\\
&= \frac{le^{-(r_i + r_j)/\lambda}}{1 + (l-1)e^{-\Delta/\lambda}}.
\end{split}
\end{equation}\qed

Moving on to the theorem, first note that it analyzes the maximum error realizable under the given constraints, not an expected error. $\mK^\textnormal{sp}$ is correct for all pairs inside a cluster and 0 otherwise. We therefore obtain the maximum error by considering the closest possible pair between clusters. By definition, this pair has distance $D-2r$ and thus
\begin{equation}
     \max_{\vx_{\textnormal{p}i}, \vx_{\textnormal{q}j}} \mK - \mK^\textnormal{sp} = e^{-(D-2r)/\lambda}
\end{equation}
LCN is also correct for all pairs inside a cluster, so we again consider the closest possible pair $\vx_i$, $\vx_j$ between clusters. We furthermore use \cref{lem:nystrombound} to only consider the landmarks of the two concerned clusters, adding an error of $\mathcal{O}(e^{-2(D-r)/\lambda})$, since $r \ll D$. Hence,
\begin{equation}
\begin{split}
    \mK_{\textnormal{LCN}, ij}^{\textnormal{2L}} &=
    \begin{pmatrix} e^{-r/\lambda} & e^{-(D-r)/\lambda} \end{pmatrix}
    \begin{pmatrix}
    1 & e^{-D/\lambda}\\
    e^{-D/\lambda} & 1
    \end{pmatrix}^{-1}
    \begin{pmatrix}
    e^{-(D-r)/\lambda}\\
    e^{-r/\lambda}
    \end{pmatrix}\\
    &= \frac{1}{1- e^{-2D/\lambda}}
    \begin{pmatrix} e^{-r/\lambda} & e^{-(D-r)/\lambda} \end{pmatrix}
    \begin{pmatrix}
    1 & -e^{-D/\lambda}\\
    -e^{-D/\lambda} & 1
    \end{pmatrix}
    \begin{pmatrix}
    e^{-(D-r)/\lambda}\\
    e^{-r/\lambda}
    \end{pmatrix}\\
    &= \frac{1}{1- e^{-2D/\lambda}}
    \begin{pmatrix} e^{-r/\lambda} & e^{-(D-r)/\lambda} \end{pmatrix}
    \begin{pmatrix}
    e^{-(D-r)/\lambda} - e^{-(D+r)/\lambda})\\
    e^{-r/\lambda} - e^{-(2D-r)/\lambda}
    \end{pmatrix}\\
    &= \frac{1}{1- e^{-2D/\lambda}}
    (e^{-D/\lambda} - e^{-(D+2r)/\lambda} + e^{-D/\lambda} - e^{-(3D-2r)/\lambda})\\
    &= \frac{e^{-D/\lambda}}{1- e^{-2D/\lambda}} (2 - e^{-2r/\lambda} - e^{-(2D-2r)/\lambda})\\
    &= e^{-D/\lambda} (2 - e^{-2r/\lambda}) - \mathcal{O}(e^{-2(D-r)/\lambda})
\end{split}
\end{equation}
and thus
\begin{equation}
\begin{split}
    \max_{\vx_{\textnormal{p}i}, \vx_{\textnormal{q}j}} \mK - \mK_\textnormal{LCN} = {}& e^{-(D-2r)/\lambda} (1 - e^{-2r/\lambda} (2 - e^{-2r/\lambda})\\
    &+ \mathcal{O}(e^{-2D/\lambda})).
\end{split}
\end{equation}
For pure Nyström we need to consider the distances inside a cluster. In the worst case two points overlap, i.e.\ $\mK_{ij} = 1$, and lie at the boundary of the cluster. Since $r \ll D$ we again use \cref{lem:nystrombound} to only consider the landmark in the concerned cluster, adding an error of $\mathcal{O}(e^{-2(D-r)/\lambda})$.
\begin{equation}
    \mK_{\textnormal{Nys}, ij} = e^{-2r/\lambda} + \mathcal{O}(e^{-2(D-r)/\lambda})
\end{equation}\qed

Note that when ignoring the effect from other clusters we can generalize the Nyström error to $l \le d$ landmarks per cluster. In this case, because of symmetry we can optimize the worst-case distance from all cluster landmarks by putting them on an $(l-1)$-simplex centered on the cluster center. Since there are at most $d$ landmarks in each cluster there is always one direction in which the worst-case points are $r$ away from all landmarks. The circumradius of an $(l-1)$-simplex with side length $\Delta$ is $\sqrt{\frac{l-1}{2l}} \Delta$. Thus, the maximum distance to all landmarks is $\sqrt{r^2 + \frac{l-1}{2l} \Delta^2}$. Using \cref{lem:nystrompair} we therefore obtain the Nyström approximation
\begin{equation}
    \mK_{\textnormal{Nys}, ij}^{\text{multi}} = \frac{l e^{-2 \sqrt{r^2 + \frac{l - 1}{2l}\Delta^2}/\lambda}}{1 + (l-1) e^{-\Delta/\lambda}} + \mathcal{O}(e^{-2(D-r)/\lambda})
\end{equation}

\section{Notes on \cref{th:sinkhorn_error}} \label{app:sinkhorn_error}

Lemmas C-F and and thus Theorem 1 by \citet{altschuler_massively_2019} are also valid for $\mQ$ outside the simplex so long as $\|\mQ\|_1 = \sum_{i,j} |\mQ_{ij}| = n$ and it only has non-negative entries. Any $\tilde{\mP}$ returned by Sinkhorn fulfills these conditions if the kernel matrix is non-negative and has support. Therefore the rounding procedure given by their Algorithm 4 is not necessary for this result. 

Furthermore, to be more consistent with \cref{th:uniform_error,th:cluster_error} we use the $L_2$ distance instead of $L_2^2$ in this theorem, which only changes the dependence on $\rho$.

\section{Notes on \cref{th:convergence}} \label{app:convergence}

To adapt Theorem 1 by \citet{dvurechensky_computational_2018} to sparse matrices (i.e.\ matrices with some $\mK_{ij} = 0$) we need to redefine
\begin{equation}
    \nu := \min_{i,j} \{\mK_{ij} | \mK_{ij} > 0\},
\end{equation}
i.e.\ take the minimum only w.r.t.\ non-zero elements in their Lemma 1. We furthermore need to consider sums exclusively over these non-zero elements instead of the full $\vone$ vector in their Lemma 1.

The Sinkhorn algorithm converges since the matrix has support \citep{sinkhorn_concerning_1967}. However, the point it converges to might not exist because we only require support, not total support. Therefore, we need to consider slightly perturbed optimal vectors for the proof, i.e.\ define a negligibly small $\tilde{\varepsilon} \ll \varepsilon, \varepsilon'$ for which $|B(u^*, v^*)\vone - r| \le \tilde{\varepsilon}$, $|B(u^*, v^*)^T\vone - c| \le \tilde{\varepsilon}$.
Support furthermore guarantees that no row or column is completely zero, thus preventing any unconstrained $u_k$ or $v_k$, and any non-converging row or column sum of $B(u_k, v_k)$. With these changes in place all proofs work the same as in the dense case.

\section{Proof of \cref{prop:derivatives}} \label{app:derivatives}

\begin{theorem}[Danskin's theorem]
    Consider a continuous function $\phi: \R^k \times Z \rightarrow \R$, with the compact set $Z \subset \R^j$. If $\phi(\vx, \vz)$ is convex in $\vx$ for every $\vz \in Z$ and $\phi(\vx, \vz)$ has a unique maximizer $\bar{\vz}$, the derivative of
    \begin{equation}
        f(\vx) = \max_{\vz \in Z} \phi(\vx, \vz)
    \end{equation}
    is given by the derivative at the maximizer, i.e.
    \begin{equation}
        \frac{\partial f}{\partial \vx} = \frac{\partial \phi(\vx, \bar{\vz})}{\partial \vx}.
    \end{equation}
\end{theorem}

We start by deriving the derivatives of the distances. To show that the Sinkhorn distance fulfills the conditions for Danskin's theorem we first identify $\vx = \mC$, $\vz = \mP$, and $\phi(\mC, \mP) = -\langle \mP, \mC \rangle_\text{F} + \lambda H(\mP)$. We next observe that the restrictions $\mP \vone_m = \vp$ and $\mP^T \vone_n = \vq$ define a compact, convex set for $\mP$. Furthermore, $\phi$ is a continuous function and linear in $\mC$, i.e.\ both convex and concave for any finite $\mP$. Finally, $\phi(\mC, \mP)$ is concave in $\mP$ since $\langle \mP, \mC \rangle_\text{F}$ is linear and $\lambda H(\mP)$ is concave. Therefore the maximizer $\bar{\mP}$ is unique and Danskin's theorem applies to the Sinkhorn distance. Using
\begin{equation}
    \begin{split}
        \frac{\partial \mC_{\text{Nys},ij}}{\partial \mU_{kl}}
        &= \frac{\partial}{\partial \mU_{kl}} \left( - \lambda \log (\sum_a \mU_{ia} \mW_{aj}) \right)\\
        &= -\lambda \delta_{ik} \frac{\mW_{lj}}{\sum_a \mU_{ia} \mW_{aj}}
        = -\lambda \delta_{ik} \frac{\mW_{lj}}{\mK_{\text{Nys},ij}},
    \end{split}
\end{equation}
\begin{equation}
    \begin{split}
        \frac{\partial \mC_{\text{Nys},ij}}{\partial \mW_{kl}}
        &= \frac{\partial}{\partial \mW_{kl}} \left( - \lambda \log (\sum_a \mU_{ia} \mW_{aj}) \right)\\
        &= -\lambda \delta_{jl} \frac{\mU_{ik}}{\sum_a \mU_{ia} \mW_{aj}}
        = -\lambda \delta_{jl} \frac{\mU_{ik}}{\mK_{\text{Nys},ij}},
    \end{split}
\end{equation}
\begin{equation}
    \begin{split}
        \frac{\bar{\mP}_{\text{Nys},ij}}{\mK_{\text{Nys},ij}}
        &= \frac{\sum_b \bar{\mP}_{U, ib} \bar{\mP}_{W, bj}}{\sum_a \mU_{ia} \mW_{aj}}
        = \frac{\bar{\vs}_i \bar{\vt}_j \sum_b \mU_{ib} \mW_{bj}}{\sum_a \mU_{ia} \mW_{aj}}\\
        &= \bar{\vs}_i \bar{\vt}_j \frac{\sum_b \mU_{ib} \mW_{bj}}{\sum_a \mU_{ia} \mW_{aj}}
        = \bar{\vs}_i \bar{\vt}_j
    \end{split}
\end{equation}
and the chain rule we can calculate the derivative w.r.t.\ the cost matrix as
\begin{equation}
    \frac{\partial d_c^\lambda}{\partial \mC}
    = -\frac{\partial }{\partial \mC} \left( -\langle \bar{\mP}, \mC \rangle_\text{F} + \lambda H(\bar{\mP}) \right)
    = \bar{\mP},
\end{equation}
\begin{equation}
    \begin{split}
        \frac{\partial d_{\text{LCN},c}^\lambda}{\partial \mU_{kl}}
        &= \sum_{i,j} \frac{\partial \mC_{\text{Nys},ij}}{\partial \mU_{kl}} \frac{\partial d_{\text{LCN},c}^\lambda}{\partial \mC_{\text{Nys},ij}}
        = -\lambda \sum_{i,j} \delta_{ik} \mW_{lj} \frac{\bar{\mP}_{\text{Nys},ij}}{\mK_{\text{Nys},ij}}\\
        &= -\lambda \sum_{i,j} \delta_{ik} \mW_{lj} \bar{\vs}_i \bar{\vt}_j
        = -\lambda \bar{\vs}_k \sum_{j} \mW_{lj} \bar{\vt}_j\\
        &= \left( -\lambda \bar{\vs} (\mW \bar{\vt})^T \right)_{kl},
    \end{split}
\end{equation}
\begin{equation}
    \begin{split}
        \frac{\partial d_{\text{LCN},c}^\lambda}{\partial \mW_{kl}}
        &= \sum_{i,j} \frac{\partial \mC_{\text{Nys},ij}}{\partial \mW_{kl}} \frac{\partial d_{\text{LCN},c}^\lambda}{\partial \mC_{\text{Nys},ij}}
        = -\lambda \sum_{i,j} \delta_{jl} \mU_{ik} \frac{\bar{\mP}_{\text{Nys},ij}}{\mK_{\text{Nys},ij}}\\
        &= -\lambda \sum_{i,j} \delta_{jl} \mU_{ik} \bar{\vs}_i \bar{\vt}_j
        = -\lambda \left( \sum_{i} \bar{\vs}_i \mU_{ik} \right) \bar{\vt}_l\\
        &= \left( -\lambda (\bar{\vs}^T \mU)^T \bar{\vt}^T \right)_{kl},
    \end{split}
\end{equation}
and $\frac{\partial d_{\text{LCN},c}^\lambda}{\partial \log \mK^\text{sp}}$ and $\frac{\partial d_{\text{LCN},c}^\lambda}{\partial \log \mK^\text{sp}_\text{Nys}}$ follow directly from $\frac{\partial d_c^\lambda}{\partial \mC}$.
We can then backpropagate in time $\mathcal{O}((n+m) l^2)$ by computing the matrix-vector multiplications in the right order.
\qed

\section{Choosing LSH neighbors and Nyström landmarks} \label{app:landmarks}

We focus on two LSH methods for obtaining near neighbors. Cross-polytope LSH \citep{andoni_practical_2015} uses a random projection matrix $\mR \in \R^{d \times b/2}$ with the number of hash buckets $b$, and then decides on the hash bucket via $h(\vx) = \argmax([\vx^T \mR \,\|\, \unaryminus \vx^T \mR])$, where $\|$ denotes concatenation. $k$-means LSH computes $k$-means and uses the clusters as hash buckets. 

We further improve the sampling probabilities of cross-polytope LSH via the AND-OR construction. In this scheme we calculate $B \cdot r$ hash functions, divided into $B$ sets (hash bands) of $r$ hash functions each. A pair of points is considered as neighbors if any hash band matches completely. $k$-means LSH does not work well with the AND-OR construction since its samples are highly correlated. For large datasets we use hierarchical $k$-means instead \citep{pauleve_locality_2010,nister_scalable_2006}.

The 3D point clouds, uniform data and the graph transport network (GTN) use the $L_2$ distance between embeddings as a cost function. For these we use (hierarchical) $k$-means LSH and $k$-means Nyström in both sparse Sinkhorn and LCN-Sinkhorn.

Word embedding similarities are measured via a dot product. In this case we use cross-polytope LSH for sparse Sinkhorn in this case. For LCN-Sinkhorn we found that using $k$-means LSH works better with Nyström using $k$-means++ sampling than cross-polytope LSH. This is most likely due to a better alignment between LSH samples and Nyström. We convert the cosine similarity to a distance via $d_{\cos} = \sqrt{1 - \frac{\vx_\text{p}^T \vx_\text{q}}{\|\vx_\text{p}\|_2 \|\vx_\text{q}\|_2}}$ \citep{berg_harmonic_1984} to use $k$-means with dot product similarity. Note that this is actually based on cosine similarity, not the dot product. Due to the balanced nature of OT we found this more sensible than maximum inner product search (MIPS). For both experiments we also experimented with uniform and recursive RLS sampling but found that the above mentioned methods work better.

\section{Implementational details} \label{app:implementation}

Our implementation runs in batches on a GPU via PyTorch \citep{paszke_pytorch_2019} and PyTorch Scatter \citep{fey_fast_2019}. To avoid over- and underflows we use log-stabilization throughout, i.e.\ we save all values in log-space and compute all matrix-vector products and additions via the log-sum-exp trick $\log \sum_i e^{x_i} = \max_j x_j + \log(\sum_i e^{x_i - \max_j x_j})$. Since the matrix $\mA$ is small we compute its inverse using double precision to improve stability. Surprisingly, we did not observe any benefit from using the Cholesky decomposition or not calculating $\mA^{-1}$ and instead solving the equation $\mB = \mA \mX$ for $\mX$. We furthermore precompute $\mW = \mA^{-1} \mV$ to avoid unnecessary operations.

We use 3 layers and an embedding size $H_\text{N} = 32$ for GTN. The MLPs use a single hidden layer, biases and LeakyReLU non-linearities. The single-head MLP uses an output size of $H_\text{N, match} = H_\text{N}$ and a hidden embedding size of $4 H_\text{N}$, i.e.\ the same as the concatenated node embedding, and the multi-head MLP uses a hidden embedding size of $H_\text{N}$. To stabilize initial training we scale the node embeddings by $\frac{\bar{d}}{\bar{n} \sqrt{H_{\text{N, match}}}}$ directly before calculating OT. $\bar{d}$ denotes the average graph distance in the training set, $\bar{n}$ the average number of nodes per graph, and $H_{\text{N, match}}$ the matching embedding size, i.e.\ \num{32} for single-head and \num{128} for multi-head OT.

For the graph datasets, the 3D point clouds and random data we use the $L_2$ distance for the cost function. For word embedding alignment we use the dot product, since this best resembles their generation procedure.

\section{Graph dataset generation and experimental details} \label{app:graph}

\begin{table*}
    \centering
    \caption{Graph dataset statistics.}
    \begin{tabular}{lcc@{\hspace{0.2cm}}cccc@{\hspace{0.2cm}}c@{\hspace{0.2cm}}c@{\hspace{0.2cm}}c}
               &                &          & \multicolumn{2}{c}{Distance (test set)} & Graphs         & Avg.\ nodes & Avg.\ edges & Node  & Edge  \\ \cline{4-5}
               & Graph type         & Distance & Mean             & Std.\ dev.            & train/val/test & per graph  & per graph  & types & types \\ \hline
AIDS30         & Molecules      & GED      & 50.5             & 16.2                 & 144/48/48      & 20.6       & 44.6       & 53    & 4     \\
Linux          & Program dependence              & GED      & 0.567            & 0.181                & 600/200/200    & 7.6        & 6.9        & 7     & -     \\
Pref. att.     & Initial attractiveness & GED      & 106.7            & 48.3                 & 144/48/48      & 20.6       & 75.4       & 6     & 4     \\
Pref. att. 200 & Initial attractiveness & PM       & 0.400            & 0.102                & 144/48/48      & 199.3      & 938.8      & 6     & -     \\
Pref. att. 2k  & Initial attractiveness & PM       & 0.359            & 0.163                & 144/48/48      & 2045.6     & 11330      & 6     & -     \\
Pref. att. 20k & Initial attractiveness & PM       & 0.363            & 0.151                & 144/48/48      & 20441      & 90412      & 6     & -
\end{tabular}

    \label{tab:datasets}
\end{table*}

\begin{table*}
    \centering
    \caption{Hyperparameters for the Linux dataset.}
    \begin{tabular}{lcc@{\hspace{0.2cm}}c@{\hspace{0.2cm}}ccc}
            & lr         & batchsize & layers & emb.\ size & $L_2$ reg. & $\lambda_\text{base}$ \\ \hline
SiamMPNN    & \num{1e-4} & 256       & 3      & 32        & \num{5e-4} & -                     \\
GMN         & \num{1e-4} & 20        & 3      & 64        & 0          & -                     \\ \hline
GTN, 1 head & 0.01       & 1000      & 3      & 32        & \num{1e-6} & 1.0                   \\
8 OT heads  & 0.01       & 1000      & 3      & 32        & \num{1e-6} & 1.0                   \\
Balanced OT & 0.01       & 1000      & 3      & 32        & \num{1e-6} & 2.0                  
\end{tabular}
    \label{tab:hparam_linux}
\end{table*}

\begin{table*}
    \centering
    \caption{Hyperparameters for the AIDS dataset.}
    \begin{tabular}{lcc@{\hspace{0.2cm}}c@{\hspace{0.2cm}}ccc}
            & lr         & batchsize & layers & emb.\ size & $L_2$ reg. & $\lambda_\text{base}$ \\ \hline
SiamMPNN    & \num{1e-4} & 256       & 3      & 32        & \num{5e-4} & -                     \\
SimGNN      & \num{1e-3} & 1         & 3      & 32        & 0.01       & -                     \\
GMN         & \num{1e-2} & 128       & 3      & 32        & 0          & -                     \\ \hline
GTN, 1 head & 0.01       & 100       & 3      & 32        & \num{5e-3} & 0.1                   \\
8 OT heads  & 0.01       & 100       & 3      & 32        & \num{5e-3} & 0.075                 \\
Balanced OT & 0.01       & 100       & 3      & 32        & \num{5e-3} & 0.1                   \\ \hline
Nyström     & 0.015      & 100       & 3      & 32        & \num{5e-3} & 0.2                   \\
Multiscale  & 0.015      & 100       & 3      & 32        & \num{5e-3} & 0.2                   \\
Sparse OT   & 0.015      & 100       & 3      & 32        & \num{5e-3} & 0.2                   \\
LCN-OT      & 0.015      & 100       & 3      & 32        & \num{5e-3} & 0.2                  
\end{tabular}
    \label{tab:hparam_aids}
\end{table*}

\begin{table*}
    \centering
    \caption{Hyperparameters for the preferential attachment GED dataset.}
    \begin{tabular}{lcc@{\hspace{0.2cm}}c@{\hspace{0.2cm}}ccc}
            & lr         & batchsize & layers & emb.\ size & $L_2$ reg. & $\lambda_\text{base}$ \\ \hline
SiamMPNN    & \num{1e-4} & 256       & 3      & 64        & \num{1e-3} & -                     \\
SimGNN      & \num{1e-3} & 4         & 3      & 32        & 0          & -                     \\
GMN         & \num{1e-4} & 20        & 3      & 64        & 0          & -                     \\ \hline
GTN, 1 head & 0.01       & 100       & 3      & 32        & \num{5e-4} & 0.2                   \\
8 OT heads  & 0.01       & 100       & 3      & 32        & \num{5e-3} & 0.075                 \\
Balanced OT & 0.01       & 100       & 3      & 32        & \num{5e-4} & 0.2                   \\ \hline
Nyström     & 0.02       & 100       & 3      & 32        & \num{5e-5} & 0.2                   \\
Multiscale  & 0.02       & 100       & 3      & 32        & \num{5e-5} & 0.2                   \\
Sparse OT   & 0.02       & 100       & 3      & 32        & \num{5e-5} & 0.2                   \\
LCN-OT      & 0.02       & 100       & 3      & 32        & \num{5e-5} & 0.2                  
\end{tabular}
    \label{tab:hparam_prefatt}
\end{table*}

The dataset statistics are summarized in \cref{tab:datasets}. Each dataset contains the distances between all graph pairs in each split, i.e.\ \num{10296} and \num{1128} distances for preferential attachment. The AIDS dataset was generated by randomly sampling graphs with at most 30 nodes from the original AIDS dataset \citep{riesen_iam_2008}. Since not all node types are present in the training set and our choice of GED is permutation-invariant w.r.t.\ types, we permuted the node types so that there are no previously unseen types in the validation and test sets. For the preferential attachment datasets we first generated 12, 4, and 4 undirected ``seed'' graphs (for train, val, and test) via the initial attractiveness model with randomly chosen parameters: 1 to 5 initial nodes, initial attractiveness of 0 to 4 and $1/2 \bar{n}$ and $3/2 \bar{n}$ total nodes, where $\bar{n}$ is the average number of nodes (\num{20}, \num{200}, \num{2000}, and \num{20000}). We then randomly label every node (and edge) in these graphs uniformly. To obtain the remaining graphs we edit the ``seed'' graphs between $\bar{n} / 40$ and $\bar{n} / 20$ times by randomly adding, type editing, or removing nodes and edges. Editing nodes and edges is 4x and adding/deleting edges 3x as likely as adding/deleting nodes. Most of these numbers were chosen arbitrarily, aiming to achieve a somewhat reasonable dataset and process. We found that the process of first generating seed graphs and subsequently editing these is crucial for obtaining meaningfully structured data to learn from. For the GED we choose an edit cost of 1 for changing a node or edge type and 2 for adding or deleting a node or an edge.

We represent node and edge types as one-hot vectors. We train all models except SiamMPNN (which uses SGD) and GTN on Linux with the Adam optimizer and mean squared error (MSE) loss for up to 300 epochs and reduce the learning rate by a factor of 10 every 100 steps. On Linux we train for up to 1000 epochs and reduce the learning rate by a factor of 2 every 100 steps. We use the parameters from the best epoch based on the validation set. We choose hyperparameters for all models using multiple steps of grid search on the validation set, see \cref{tab:hparam_linux,tab:hparam_aids,tab:hparam_prefatt} for the final values. We use the originally published result of SimGNN on Linux and thus don't provide its hyperparameters. GTN uses 500 Sinkhorn iterations. We obtain the final entropy regularization parameter from $\lambda_\text{base}$ via $\lambda = \lambda_\text{base} \frac{\bar{d}}{\bar{n}} \frac{1}{\log n}$, where $\bar{d}$ denotes the average graph distance and $\bar{n}$ the average number of nodes per graph in the training set. The factor $\bar{d} / \bar{n}$ serves to estimate the embedding distance scale and $1 / \log n$ counteracts the entropy scaling with $n \log n$. Note that the entropy regularization parameter was small, but always far from 0, which shows that entropy regularization actually has a positive effect on learning. On the pref.\ att.\ 200 dataset we use no $L_2$ regularization, $\lambda_\text{base} = 0.5$, and a batch size of 200. For pref.\ att.\ 2k we use $\lambda_\text{base} = 2$ and a batch size of 20 for full Sinkhorn and 100 for LCN-Sinkhorn. For pref.\ att.\ 20k we use $\lambda_\text{base} = 50$ and a batch size of 4. $\lambda_\text{base}$ scales with graph size due to normalization of the PM kernel.

For LCN-Sinkhorn we use roughly 10 neighbors for LSH (20 $k$-means clusters) and 10 $k$-means landmarks for Nyström on pref.\ att.\ 200. We double these numbers for pure Nyström Sinkhorn, sparse Sinkhorn, and multiscale OT. For pref.\ att.\ 2k we use around 15 neighbors ($10 \cdot 20$ hierarchical clusters) and 15 landmarks and for pref.\ att.\ 20k we use roughly 30 neighbors ($10 \cdot 10 \cdot 10$ hierarchical clusters) and 20 landmarks. The number of neighbors for the 20k dataset is higher and strongly varies per iteration due to the unbalanced nature of hierarchical $k$-means. This increase in neighbors and landmarks and PyTorch's missing support for ragged tensors largely explains LCN-Sinkhorn's deviation from perfectly linear runtime scaling.

We perform all runtime measurements on a compute node using one Nvidia GeForce GTX 1080 Ti, two Intel Xeon E5-2630 v4, and 256GB RAM.

\section{Runtimes} \label{app:runtimes}

\begin{figure}[t]
    \centering
    \input{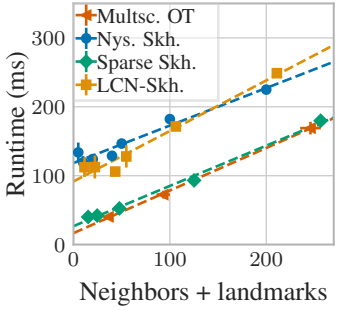}
    \caption{Runtime scales linearly with the number of neighbors/landmarks for all relevant Sinkhorn approximation methods.}
    \label{fig:runtime_neighbors}
\end{figure}

\cref{tab:runtimes} compares the runtime of the full Sinkhorn distance with different approximation methods using 40 neighbors/landmarks. We separate the computation of approximate $\mK$ from the optimal transport computation (Sinkhorn iterations), since the former primarily depends on the LSH and Nyström methods we choose. We observe a 2-4x speed difference between sparse (multiscale OT and sparse Sinkhorn) and low-rank approximations (Nyström Sinkhorn and LCN-Sinkhorn), while factored OT is multiple times slower due to its iterative refinement scheme. In \cref{fig:runtime_neighbors} we observe that this runtime gap stays constant independent of the number of neighbors/landmarks, i.e.\ the relative difference decreases as we increase the number of neighbors/landmarks. This gap could either be due to details in low-level CUDA implementations and hardware or the fact that low-rank approximations require 2x as many multiplications for the same number of neighbors/landmarks. In either case, both \cref{tab:runtimes} and \cref{fig:runtime_neighbors} show that the runtimes of all approximations scale linearly both in the dataset size and the number of neighbors and landmarks, while full Sinkhorn scales quadratically.

\begin{figure}[t]
    \centering
    \input{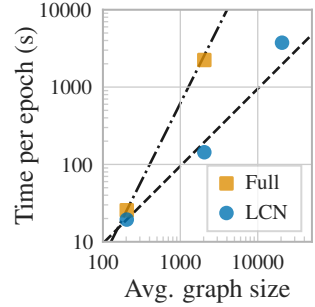}
    \caption{Log-log runtime per epoch for GTN with full Sinkhorn and LCN-Sinkhorn. LCN-Sinkhorn scales almost linearly with graph size while sustaining similar accuracy.}
    \label{fig:gtn_runtime}
\end{figure}

We furthermore investigate whether GTN with approximate Sinkhorn indeed scales log-linearly with the graph size by generating preferential attachment graphs with \num{200}, \num{2000}, and \num{20000} nodes (\SI{\pm 50}{\percent}). We use the Pyramid matching (PM) kernel \citep{nikolentzos_matching_2017} as prediction target. \cref{fig:gtn_runtime} shows that the runtime of LCN-Sinkhorn scales almost linearly (dashed line) and regular full Sinkhorn quadraticly (dash-dotted line) with the number of nodes, despite both achieving similar accuracy and LCN using slightly more neighbors and landmarks on larger graphs to sustain good accuracy. Full Sinkhorn went out of memory for the largest graphs.

\begin{table*}
    \centering
    \caption{Runtimes (ms) of Sinkhorn approximations for EN-DE embeddings at different dataset sizes. Full Sinkhorn scales quadratically, while all approximationes  scale at most linearly with the size. Sparse approximations are 2-4x faster than low-rank approximations, and factored OT is multiple times slower due to its iterative refinement scheme. Note that similarity matrix computation time ($\mK$) primarily depends on the LSH/Nyström method, not the OT approximation.}
    \begin{tabular}{l@{\hspace{0.2cm}}S[table-format=3]S[table-format=4]cS[table-format=3]S[table-format=5]cS[table-format=3]S[table-format=4]}
              & \multicolumn{2}{c}{$N=10000$} &  & \multicolumn{2}{c}{$N=20000$} &  & \multicolumn{2}{c}{$N=50000$} \\ \cline{2-3} \cline{5-6} \cline{8-9} 
              & $\mK$       & OT               &  & $\mK$       & OT               &  & $\mK$           & OT           \\ \hline
Full Sinkhorn & 8          & 2950       &  & 29         & 11760      &  & OOM            & OOM          \\
Factored OT   & 29         & 809              &  & 32         & 1016             &  & 55             & 3673         \\
Multiscale OT & 90         & 48               &  & 193        & 61               &  & 521            & 126          \\
Nyström Skh.  & 29         & 135              &  & 41         & 281              &  & 79             & 683          \\
Sparse Skh.   & 42         & 46               &  & 84         & 68               &  & 220            & 137          \\
LCN-Sinkhorn  & 101        & 116              &  & 242        & 205              &  & 642            & 624         
\end{tabular}
    \label{tab:runtimes}
\end{table*}

\section{Distance approximation} \label{app:distance}

\begin{figure*}
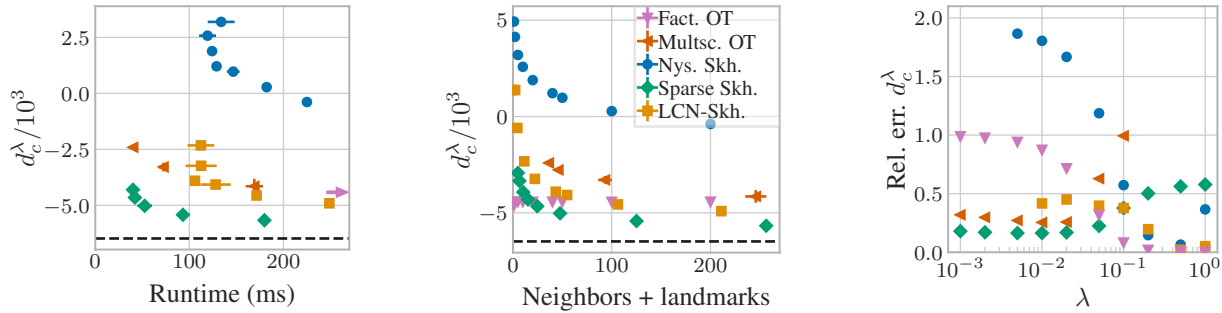

    \centering
    \begin{minipage}[t]{0.325\textwidth}
        \input{figures/dist_runtime.pgf}
    \end{minipage}
    \hfill
    \begin{minipage}[t]{0.325\textwidth}
        \input{figures/dist_neighbors.pgf}
    \end{minipage}
    \hfill
    \begin{minipage}[t]{0.325\textwidth}
        \input{figures/rel_dist_reg.pgf}
    \end{minipage}
    \caption{Sinkhorn distance approximation for different runtimes and computational budgets (both varied via the number of neighbors/landmarks), and entropy regularization parameters $\lambda$. The dashed line denotes the true Sinkhorn distance. The arrow indicates factored OT results far outside the depicted range. \textbf{Left:} Sparse Sinkhorn consistently performs best across all runtimes. \textbf{Center:} Sparse Sinkhorn mostly performs best, with LCN-Sinkhorn coming in second, and factored OT being seemingly independent from the number of neighbors. \textbf{Right:} Sparse Sinkhorn performs best for low $\lambda$, LCN-Sinkhorn for moderate and high $\lambda$ and factored OT for very high $\lambda$.}
    \label{fig:dist}
\end{figure*}

\cref{fig:dist} shows that for the chosen $\lambda = 0.05$ sparse Sinkhorn offers the best trade-off between computational budget and distance approximation, with LCN-Sinkhorn and multiscale OT coming in second. Factored OT is again multiple times slower than the other methods. Note that $d_c^\lambda$ can be negative due to the entropy offset. This picture changes as we increase the regularization. For higher regularizations LCN-Sinkhorn is the most precise at constant computational budget (number of neighbors/landmarks). Note that the crossover points in this figure roughly coincide with those in \cref{fig:pcc}. Keep in mind that usually the OT plan is more important than the raw distance approximation, since it determines the training gradient and tasks like embedding alignment don't use the distance at all. This becomes evident in the fact that sparse Sinkhorn achieves a better distance approximation than LCN-Sinkhorn but performs worse in both downstream tasks investigated in \cref{sec:exp}.

\end{document}